\documentclass[10pt,journal,compsoc]{IEEEtran}

\ifCLASSOPTIONcompsoc
  \usepackage[nocompress]{cite}
\else
  \usepackage{cite}
\fi

\ifCLASSINFOpdf
\else
\fi

\usepackage{amsmath,amsfonts}
\usepackage{algorithmic}
\usepackage{algorithm}
\usepackage{array}
\usepackage[caption=false,font=normalsize,labelfont=sf,textfont=sf]{subfig}
\usepackage{textcomp}
\usepackage{stfloats}
\usepackage{url}
\usepackage{verbatim}
\usepackage{graphicx}
\usepackage{xcolor}
\usepackage{bm}
\usepackage{bbding}
\usepackage{pifont}
\usepackage{makecell}
\usepackage{multirow}
\usepackage{mathtools}

\begin{document}
\title{SpotFormer: Multi-Scale Spatio-Temporal Transformer for Facial Expression Spotting}

\author{Yicheng~Deng, 
Hideaki~Hayashi,~\IEEEmembership{Member,~IEEE,}
and~Hajime~Nagahara,~\IEEEmembership{Member,~IEEE}

\IEEEcompsocitemizethanks{\IEEEcompsocthanksitem Y. Deng is with the Graduate School of Information Science and Technology, The University of Osaka, Suita, 565-0871, Japan. Email: yicheng@is.ids.osaka-u.ac.jp\protect\\
\IEEEcompsocthanksitem H. Hayashi and H. Nagahara are with the D3 Center, The University of Osaka, Suita, 565-0871, Japan. Email: hayashi@ids.osaka-u.ac.jp; nagahara@ids.osaka-u.ac.jp}
\thanks{Manuscript received April 19, 2005; revised August 26, 2015. (Corresponding author: Yicheng Deng.)}}

\markboth{Journal of \LaTeX\ Class Files,~Vol.~14, No.~8, August~2015}%
{Shell \MakeLowercase{\textit{et al.}}: Bare Demo of IEEEtran.cls for Computer Society Journals}


\IEEEtitleabstractindextext{%
\begin{abstract}
Facial expression spotting, identifying periods where facial expressions occur in a video, is a significant yet challenging task in facial expression analysis. The issues of irrelevant facial movements and the challenge of detecting subtle motions in micro-expressions remain unresolved, hindering accurate expression spotting. In this paper, we propose an efficient framework for facial expression spotting. First, we propose a Compact Sliding-Window-based Multi-temporal-Resolution Optical flow (CSW-MRO) feature, which calculates multi-temporal-resolution optical flow of the input image sequence within compact sliding windows. The window length is tailored to perceive complete micro-expressions and distinguish between general macro- and micro-expressions. CSW-MRO can effectively reveal subtle motions while avoiding the optical flow being dominated by head movements. Second, we propose SpotFormer, a multi-scale spatio-temporal Transformer that simultaneously encodes spatio-temporal relationships of the CSW-MRO features for accurate frame-level probability estimation. In SpotFormer, we use the proposed Facial Local Graph Pooling (FLGP) operation and convolutional layers to extract multi-scale spatio-temporal features. We show the validity of the architecture of SpotFormer by comparing it with several model variants. Third, we introduce supervised contrastive learning into SpotFormer to enhance the discriminability between different types of expressions. Extensive experiments on SAMM-LV, CAS(ME)$^\mathrm{2}$, and CAS(ME)$^\mathrm{3}$ show that our method outperforms state-of-the-art models, particularly in micro-expression spotting. Code is available at \url{https://github.com/KinopioIsAllIn/SpotFormer}.
\end{abstract}

}

\maketitle

\IEEEdisplaynontitleabstractindextext

%
\IEEEpeerreviewmaketitle

\section{Introduction}
\label{intro}
\IEEEPARstart{F}{acial} expressions are a fundamental aspect of nonverbal communication and play a crucial role in conveying human emotions.
As people experience emotional changes, facial muscles undergo voluntary or involuntary movements, resulting in various expressions. These expressions act as powerful and direct social signals, enabling others to understand their emotions and enhancing interpersonal communication.

Facial expressions can be broadly categorized into two groups: macro-expressions (MaEs) and micro-expressions (MEs). MaEs typically last from 0.5 to 4.0 seconds~\cite{nummenmaaekman} and are easily perceived by people due to their occurrence on a large facial area and high intensity \cite{corneanu2016survey}. The analysis of MaEs is significant in various practical applications, such as sociable robots \cite{fukuda2002facial}, mental health \cite{kopper1996experience}, and virtual reality \cite{facialvr}. In contrast, MEs generally last for less than 0.5 seconds \cite{yan2013fast} (mostly distributed around 306–328 ms~\cite{li2025could}), and their perception is much more challenging due to their localized occurrence~\cite{bhushan2015study} and low intensity \cite{wang2025micro}. Because of their involuntary nature, MEs are crucial in situations where people may attempt to conceal emotions or deceive others, such as in lie detection \cite{ekman2009telling}, medical care \cite{endres2009micro}, and national security \cite{o2009police}. Therefore, both MaE and ME analysis play important roles in understanding human emotions and behaviors.

\begin{figure}[t]
\centering
\includegraphics[width=0.48\textwidth]{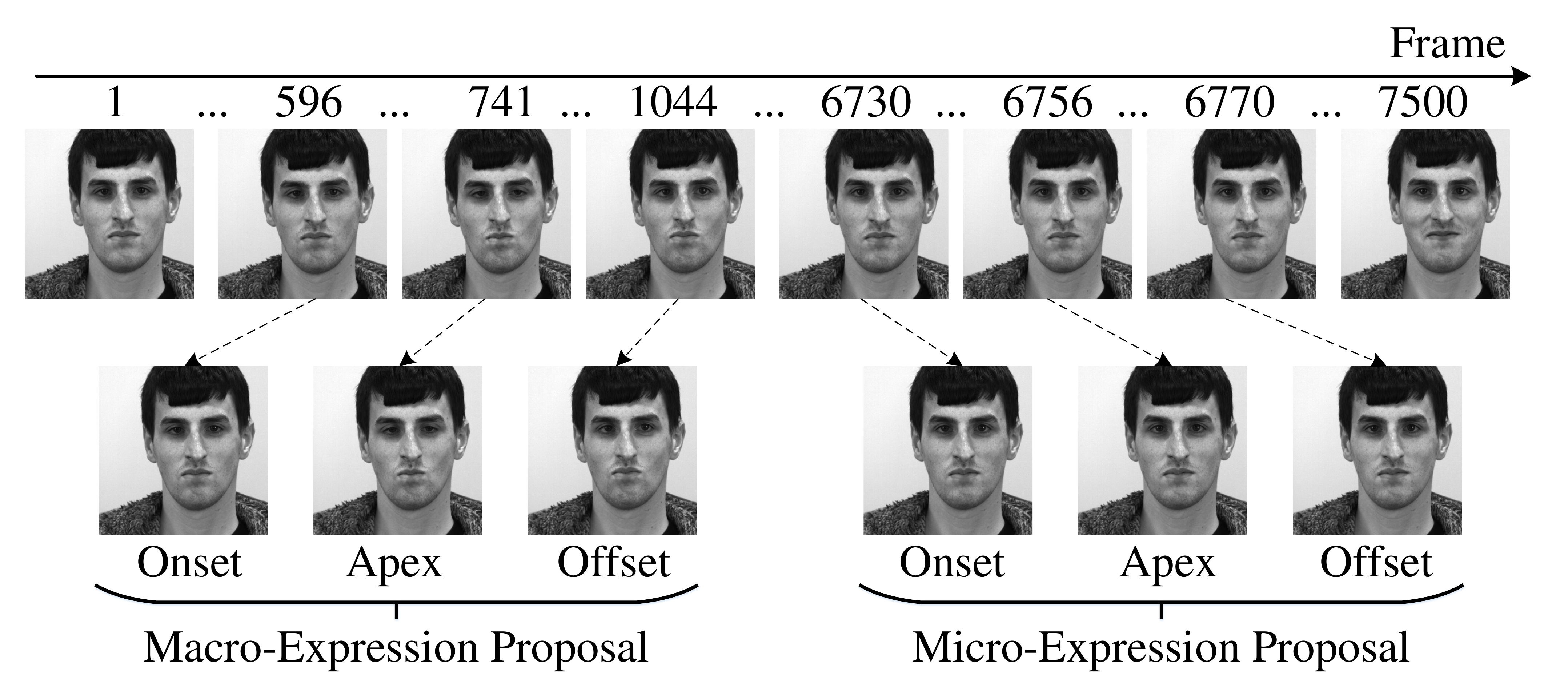}
\caption{Illustration of macro- and micro-expression spotting.}
\label{abstract}
\end{figure}

Facial expression analysis generally consists of two steps: facial expression spotting (FES) and facial expression recognition~\cite{zhang2017facial, wang2025visual}.
As the initial phase, FES aims to locate the onset and offset frames of MaE and ME intervals within video sequences, as illustrated in Fig.~\ref{abstract}. The onset frame represents the beginning of an expression, while the offset frame indicates its end. However, FES is highly challenging due to several factors, including the difficulty in detecting subtle motions that exist in MEs, the presence of irrelevant motions (e.g., head movements and eye blinking) in long videos, and the difficulty of distinguishing between MaEs, MEs, and neutral expressions.

In recent years, researchers have made significant progress in developing efficient algorithms for FES. Early studies employed traditional methods to extract hand-crafted features, such as optical flow in large sliding windows. Subsequently, they analyzed feature variations and detected expression intervals using threshold strategies \cite{gan2020optical, he2020spotting, yuhong2021research, zhao2022rethinking}. More recently, with the development of deep learning, more and more researchers introduced learning-based methods to address the problem \cite{verburg2019micro, wang2021mesnet, yap20223d, yin2023aware}. Additionally, due to the similarity between FES and temporal action localization (TAL), some researchers applied classical TAL frameworks to FES\cite{yu2023lgsnet, GUO2023146} and introduced additional modules. These methods have achieved considerable improvements in MaE spotting performance.

Despite significant progress made by previous studies, there are still some open questions that require exploration for further performance improvement, especially for ME spotting.
\textbf{The first question} is how to extract more robust motion features. Recent traditional methods typically employ a large sliding window strategy to extract optical flow features and spot potential expression proposals within each sliding window \cite{he2020spotting, guo2021magnitude}. However, the accuracy of using such a strategy is significantly impacted by head movement issues. In contrast, recent deep learning methods that employ optical flow for temporal modeling calculate optical flow between adjacent frames as motion features \cite{yu2021lssnet, leng2022abpn}. Nevertheless, such an optical flow extraction method is insufficient to effectively reveal subtle motions present in MEs. Therefore, there is an urgent need to develop a strategy that can magnify motion information while mitigating the influence of head movement.
\textbf{Second}, some methods \cite{GUO2023146, yu2023lgsnet} applied general TAL frameworks to FES and refined them by introducing novel modules. However, these methods usually employ a pre-trained I3D\cite{carreira2017quo} network to extract features from snippets based on the optical flow extracted between adjacent frames, which is unsuitable for revealing subtle motions as we mentioned before. Consequently, their performance, particularly in ME spotting, requires improvement.
\textbf{Third}, some methods compute optical flow in specific regions of interest (ROIs) where facial expressions frequently happen to alleviate the influence of irrelevant motions. They then design a network to learn the extracted motion features \cite{leng2022abpn, yin2023aware}. However, their proposed networks need a more comprehensive consideration of spatio-temporal relationships and multi-scale feature learning. This limitation restricts the representational capacity of the models.

To address the issues mentioned above, we propose an efficient framework for facial expression spotting in untrimmed videos. \textbf{First}, we propose a \textit{compact sliding-window-based multi-temporal-resolution} optical flow (CSW-MRO) feature to amplify motion information while mitigating head movement problems. Specifically, we calculate the optical flow between the first frame and each subsequent frame within compact sliding windows to generate multi-temporal-resolution optical flow features. Moreover, the temporal sliding window size is tailored to perceive complete subtle MEs and discern the differences between general MEs and MaEs. The CSW-MRO feature enables our framework to achieve accurate frame-level apex or boundary (onset or offset) probability estimation and expression type classification.
\textbf{Second}, we propose SpotFormer, a multi-scale spatio-temporal transformer to capture multi-scale spatial relationships and temporal variations among different facial parts across frames from the CSW-MRO features. In SpotFormer, a facial local graph pooling (FLGP) operation is designed to extract multi-scale facial spatial features, while a learning-based temporal downsampling strategy is employed to extract multi-scale temporal features. These designs enhance the model’s understanding ability from low-level to high-level. 
The exploration of several model variants and extensive comparative experiments validate the effectiveness of SpotFormer. \textbf{Third}, we introduce supervised contrastive learning into our model to learn a finer discriminative feature representation. This contrastive learning approach alleviates the challenge of recognizing the classification boundaries between MaEs, MEs, and neutral expressions at the frame level.

This paper is an extension of our prior conference publication \cite{deng2024SpoTGCN}. The most crucial update is the proposal of SpotFormer. As other differences from the conference paper version, we make a more extensive overview and comparison of the related literature, particularly with regard to TAL. In addition, we present a more comprehensive qualitative and quantitative comparison to previous work in the experimental section. The contributions are listed as follows, where * represents new contributions of this paper:
\begin{itemize}
\item We propose CSW-MRO, a compact sliding-window-based multi-temporal-resolution optical flow feature for facial expression spotting, which can magnify the motion information while avoiding the optical flow being dominated by head movements. Extensive ablation studies demonstrate that such a carefully designed optical flow feature can assist the model not only in perceiving complete subtle MEs but also in discerning the differences between general MaEs and MEs.
\item * We present SpotFormer, a novel Transformer-based framework to capture spatial relationships and temporal variations at multiple scales for accurate frame-level probability estimation. In SpotFormer, we propose a facial local graph pooling operation tailored to the facial graph structure for multi-scale spatial feature learning, while learnable convolution layers are employed for multi-scale temporal feature learning. Extensive comparative experiments on various model architectures verify the effectiveness of SpotFormer.
\item We introduce supervised contrastive learning into our model to enhance the discriminative feature representation. To the best of our knowledge, our work is the first to study contrastive learning for facial expression spotting, which enables our model to learn better classification boundaries between different expression categories at the frame level.
\end{itemize}

\section{Related Work}
\label{rela}
\subsection{Facial expression spotting}
Early FES methods extracted appearance-based features, such as local binary patterns \cite{moilanen2014spotting} and histogram of oriented gradients \cite{davison2015micro}. These methods then employed machine learning algorithms and threshold strategies for expression spotting.
Subsequently, See et al. \cite{see2019megc} proposed the Facial Micro-Expressions Grand Challenge (MEGC), which introduced unified evaluation datasets and metrics for the FES task. The subsequent MEGC series \cite{jingting2020megc2020, li2022megc2022, davison2023megc2023, see2024megc2024} have attracted increasing attention from researchers and significantly promoted the development of this field. 

In recent years, the mainstream approach for motion feature extraction shifted to optical flow. He et al. \cite{he2020spotting} employed main directional maximal difference analysis \cite{wang2017main} to detect facial movements and spot potential expression proposals.
He~\cite{yuhong2021research} alleviated head movement problems by repeating face alignment using optical flow in the nose tip region. Guo et al.~\cite{guo2021magnitude} converted optical flow vectors into a polar coordinate representation and introduced a novel decision criterion accordingly. Recently, several researchers \cite{zhao2022rethinking, yu2024micro} proposed novel methods to refine optical flow features or expression proposals, which further improved performance.

With the development of deep learning, an increasing number of researchers have proposed various neural networks for feature learning.
Zhang et al. \cite{zhang2018smeconvnet} employed convolutional neural networks (CNNs) to extract features from video clips and spotted apex frames by analyzing the feature representations.
Verburg et al. \cite{verburg2019micro} extracted histogram of oriented optical flow features and employed an RNN to spot expression intervals. 
Liong et al. \cite{liong2021shallow} introduced a three-stream CNN and employed pseudo-labeling techniques to facilitate the learning process.
Yang et al. \cite{yang2021facial} incorporated facial action unit information and concatenated various types of neural networks for feature learning.
Leng et al. \cite{leng2022abpn} calculated the main directional mean optical flow (MDMO) \cite{liu2015main} optical flow features between adjacent frames and extended BSN~\cite{Lin_2018_ECCV} for frame-level probability estimation.
Yin et al. \cite{yin2023aware} extended ABPN~\cite{leng2022abpn} by introducing a graph convolutional network (GCN) to embed action unit (AU) label information into the extracted optical flow. 
Guo et al. \cite{GUO2023146} proposed a multi-scale local transformer, similar to ActionFormer \cite{zhang2022actionformer}, and considered multi-scale temporal feature fusion for performance improvement.
Yu et al. \cite{yu2023lgsnet} applied A2Net \cite{yang2020revisiting} to facial expression spotting and improved it by introducing attention modules. Zhang et al.~\cite{zhang2024multi} proposed a framework that integrates VideoMAE-based feature extraction, multi-scale candidate segment generation, and a multi-start-point optical flow filtering method for accurate ME and MaE detection. Zou et al.~\cite{zou2025synergistic} proposed a temporal state transition architecture based on state space models, which formulates spotting and recognition as a time-series regression problem rather than window-level classification.

While current optical flow-based methods have significantly improved MaE spotting, their performance in ME spotting remains considerably lower. This is attributed to their extracted optical flow features, which cannot efficiently reveal subtle motions that exist in MEs. To solve this issue, we propose to extract a compact sliding-window-based multi-temporal-resolution optical flow feature, which can effectively magnify these subtle motions and enhance the distinction between MaEs and MEs, thereby achieving accurate frame-level probability estimation and improving MaE and ME spotting performance.

\subsection{Temporal action localization}
Similar to FES, TAL aims to detect action intervals in general scene videos.
Early works treated TAL as a 1D object detection task. Xu et al. \cite{Xu_2017_ICCV} proposed a region convolutional 3D network to encode the video streams and locate the action proposals using an anchor-based method. Chao et al. \cite{Chao_2018_CVPR} applied the faster R-CNN object detection framework to TAL by employing a multi-scale architecture to accommodate extreme variations in action duration. However, anchor-based methods require pre-defining a fixed number of anchors. The length and number of anchors need careful pre-definition, limiting flexibility and affecting the generalization ability of the proposed model.

In 2018, Lin et al. \cite{Lin_2018_ECCV} proposed the Boundary-Sensitive Network (BSN), which was the first method to estimate frame-level probabilities and subsequently became the mainstream approach.
Recently, Nag et al. \cite{Nag_2023_CVPR} presented a novel Gaussian aware post-processing method for more accurate model inference, which models the start and end points of action instances with a Gaussian distribution to enable temporal boundary inference at a sub-snippet level. 

Yang et al. \cite{yang2020revisiting} were the first to directly regress the boundary offset for each temporal point, which solves the inflexibility and computation problems of anchor-based methods. Zhang et al. presented ActionFormer \cite{zhang2022actionformer}, a simple yet effective multi-scale transformer designed to further improve action localization performance. 

Even though there are many similarities between TAL and FES, there are still some differences between the two tasks that may limit the performance improvements when \textbf{directly} applying TAL methods to FES.

Here, we summarize three main differences that motivate us to consider the specific characteristics of the FES task more thoroughly when developing our unique method.

Firstly, motion information is usually obvious to detect in TAL videos, while in FES, the main challenge is to capture the subtle motion of MEs and suppress noises such as head movement and eye blinking. Since anchor-based methods and boundary regression-based methods typically require a large receptive field, as discussed before, extracting delicate optical flow becomes challenging. Therefore, in this paper, we design a frame-level probability-based framework following BSN \cite{Lin_2018_ECCV}.

Secondly, the scene usually changes in TAL videos, and the main challenge is to distinguish between the action scene and background change. In contrast, the scene during the whole video does not change much in FES, making it easier to capture vital motion information in certain facial areas that are strongly related to facial expressions. Therefore, in this paper, we represent the CSW-MRO feature as spatio-temporal facial graphs and propose the SpotFormer to efficiently analyze the latent motion patterns.

Finally, TAL involves an action classification task for each proposal since their videos may involve hundreds of action classes. However, there are only two action classes (i.e., MaE and ME) in FES. Rather than focusing on action recognition, we aim to distinguish between MEs and MaEs. Therefore, in this paper, we emphasize perceiving MEs and analyzing the differences between MEs and MaEs within compact sliding windows, which inspired us to propose the CSW-MRO feature and the supervised contrastive learning.

\begin{figure*}[tbp]
\centering
\includegraphics[width=1.0\textwidth]{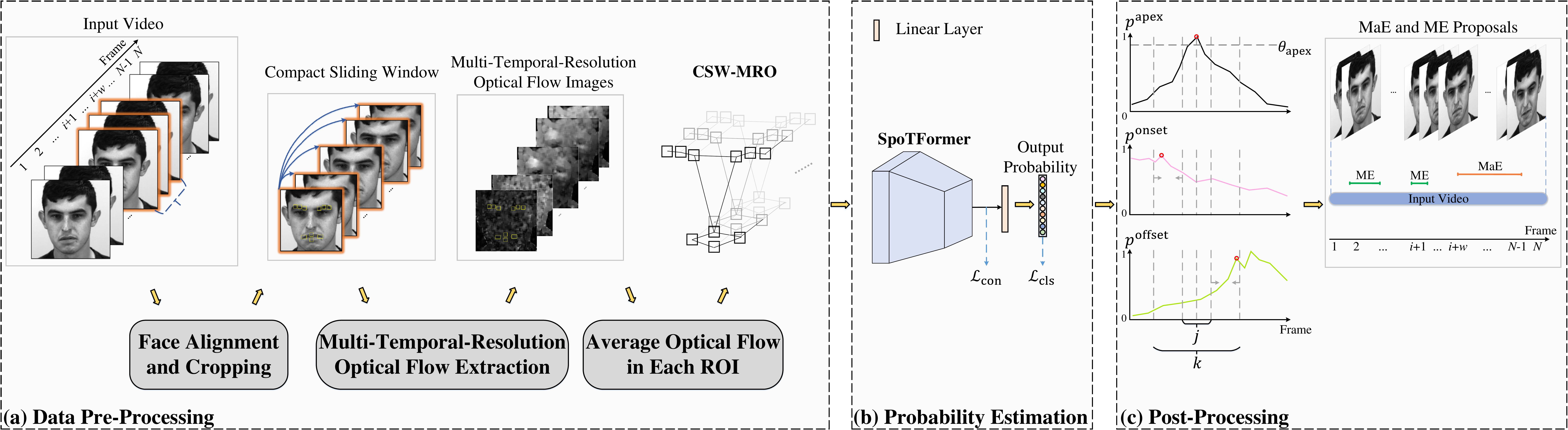}
\caption{Overview of the proposed framework. 
(a) The data pre-processing module calculates the CSW-MRO features;
(b) the probability estimation module employs SpotFormer which takes optical flow features as input for frame-level apex or boundary probability estimation; (c) the post-processing module detects pseudo-apex frames, generates expression proposals by searching for onset and offset frames around each pseudo-apex, and filters redundant proposals using NMS.
}
\label{framework}
\end{figure*}

\begin{figure}[t]
\centering
\includegraphics[width=0.23\textwidth]{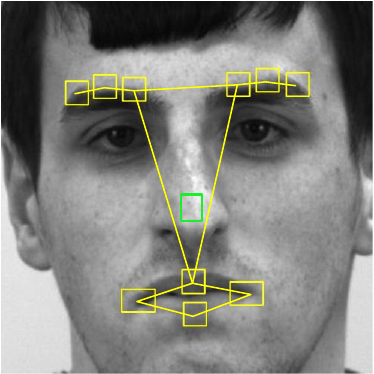}
\caption{Extracted ROIs and constructed facial graph structure are denoted in yellow, while the nose tip region for face alignment is denoted in green. 
}
\label{rois}
\end{figure}

\subsection{Transformer}
The Transformer was first proposed by Vaswani et al. \cite{transformer} for natural language processing. In recent years, many researchers have explored the Transformer for computer vision and have shown great success in 2D image processing (e.g., ViT \cite{dosovitskiy2021an}, Swin Transformer \cite{liu2021swin}) and 3D video understanding (e.g., ViViT \cite{Arnab_2021_ICCV}, Video Swin Transformer \cite{Liu_2022_CVPR}, ActionFormer \cite{zhang2022actionformer}). In recent years, several efforts have been devoted to investigating Transformers for macro- or micro-expression recognition~\cite{li2024fer, wang2024htnet}.

Such success is attributed to their powerful self-attention mechanism and ability to model long-range relations. The self-attention mechanism allows the model to focus on critical features and ignore irrelevant information, enhancing performance and robustness. The long-range relation modeling enables efficient capture of context information across both spatial and temporal dimensions, contributing to a better understanding of the overall sequence. In this paper, we explore the potential of the graph-based Transformer for facial expression spotting and introduce SpotFormer, a novel model designed to enhance the accuracy of expression spotting.

\subsection{Contrastive learning}
In recent years, contrastive learning has proven to be effective in various domains, including computer vision and natural language processing. The objective of contrastive learning in unsupervised learning \cite{chen2020simple} is to acquire discriminative representations by maximizing the similarity between similar instances (positive pairs generated using data augmentation) while minimizing the similarity between dissimilar instances (negative pairs). This approach produces meaningful visual representations applicable to tasks such as image recognition, representation learning, and semantic understanding.
Based on the idea that positive pairs can also be selected by ground-truth labels, Khosla et al. \cite{khosla2020supervised} introduced the supervised contrastive loss, showcasing its potential to enhance supervised tasks by incorporating labeled information during the training process. Supervised contrastive learning efficiently enlarges the domain discrepancy, leading to an enhanced extraction of discriminative feature representations. Our method focuses on recognizing the boundary between different types of expressions at the frame level. To achieve this objective, we establish contrasts between MaE and ME frames, as well as between ME frames and neutral frames. This approach enables our model to acquire more discriminative feature representations, ultimately reducing the misclassification rate.

\section{Proposed Expression Spotting Framework}
An overview of the proposed framework is shown in Fig. \ref{framework}. Given a raw video as input, the proposed framework aims to detect all potential MaE and ME intervals within the video, spotting the onset and offset frames, as well as assigning the expression type for each interval. The framework consists of three modules: the data pre-processing module, the probability estimation module, and the post-processing module.

\begin{figure*}[t]
\centering
\includegraphics[width=1.0\textwidth]{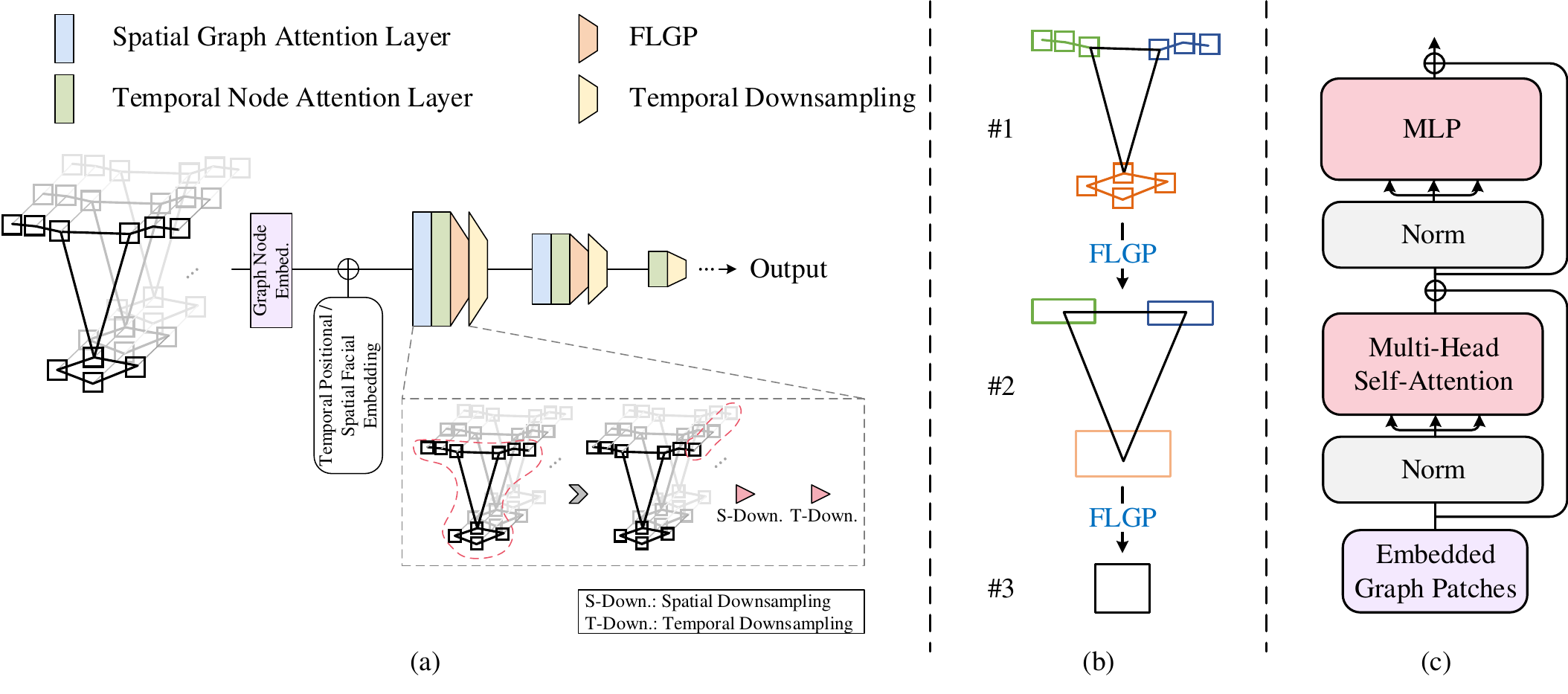}
\caption{Overview of the proposed SpotFormer. (a) The network structure of SpotFormer. The dotted gray box provides an intuitive illustration of the spatio-temporal attention block. A spatial graph attention layer models the spatial relationships within the graph of each frame, while a temporal node attention layer captures temporal variations of each graph node across frames. Subsequently, spatial and temporal downsampling are performed using FLGP and a learnable temporal downsampling layer, respectively. (b) The scale change between different facial graph structures through FLGP. Each facial muscle is represented by a distinct color (e.g., orange represents the lips). In the first spatial downsampling, ROIs belonging to the same facial muscle are aggregated into single nodes via FLGP to construct graph \#2. In the second spatial downsampling, all aggregated nodes are further merged into a single node via FLGP to form graph \#3. (c) The illustration of the spatial graph attention layer and temporal node attention layer.}
\label{model1}
\end{figure*}

\subsection{Data pre-processing}
Assuming that a video $V=(v_i)_{i=1}^{N}$ with $N$ frames, we initially calculate the CSW-MRO features.
First, by setting a window length $w$, we pad the beginning and end of the video with $\lfloor \frac{w}{2} \rfloor$ repetitions of the first and last frames, respectively. This ensures that the original first and last frames can be the center frames of the first and last sliding windows, respectively. Then, we divide the entire video into $N$ overlapped clips $C=(c_i)_{i=1}^N$ with the window length $w$ and sliding stride 1, where $c_i$ involves $v_i$ as the center frame. 
In this paper, we aim to achieve accurate frame-level probability estimation based on compact sliding windows.
Specifically, we utilize all the temporal information of $c_i$ and magnify the motion information existing in $c_i$ to estimate the probability map of $v_i$. Next, we will introduce the process of obtaining multi-temporal-resolution optical flow features in each compact sliding window.

For each clip $c_i$, we initially detect the 68 facial keypoints using a pre-trained MobileFaceNet~\cite{PFL} in the first frame $c_i^1$. These keypoints will subsequently be used to generate the facial bounding box, extract the nose tip area for facial alignment, and extract ROIs. Specifically, we detect the facial bounding box in the first frame $c_i^1$. In each subsequent frame $c_i^s$, where $s=2, 3, \ldots, w$, we initialize the facial bounding box of $c_i^s$ with the one from $c_i^1$. Given that the nose tip area remains stationary during facial expressions~\cite{yuhong2021research}, we compute the optical flow of the nose tip area, as illustrated in Fig.~\ref{rois}, to represent the global head movement from $c_i^1$ to $c_i^s$. 
To achieve this, we employ the Farneback algorithm~\cite{farneback2003two} to compute MDMO optical flow~\cite{liu2015main} of the nose tip region from $c_i^1$ to $c_i^s$.  
The resulting feature $o_i^{s,\mathrm{nose}} \in \mathbb{R}^2$ is obtained by averaging the optical flow vectors within the extracted nose region $M_i^{\mathrm{nose}}$ that align with the main direction of optical flow:
\begin{equation}
    o_i^{s,\mathrm{nose}} = \frac{1}{n_i^{\mathrm{nose}}} \sum_{(x,y) \in M_i^{\mathrm{nose}}} \mathrm{MDMO}(c_i^1, c_i^s)[x,y],
\label{of}
\end{equation}
where $n_i^{\mathrm{nose}}$ denotes the number of pixels in $M_i^{\mathrm{nose}}$ whose motion directions are consistent with the main optical flow direction, 
and $\mathrm{MDMO}(c_i^1, c_i^s)[x,y]$ represents the MDMO optical flow vector at location $(x,y)$. 
Optical flow vectors that do not follow the main direction are suppressed by setting them to $\vec{0}$.

Consequently, $o^{s,\mathrm{nose}}_i$ contains the horizontal and vertical movements of the head from $c_i^1$ to $c_i^s$. Then, we translate the facial bounding box of frame $c_i^s$ by adding $o^{s,\mathrm{nose}}_i$ to complete facial alignment.

Based on the discussion in Sections~\ref{intro} and \ref{rela}, we selectively extract $R$ ROIs \cite{zhao2022rethinking}, where facial expressions happen most frequently, rather than processing entire images. This choice can effectively mitigate the impact of noise and irrelevant facial movements. Then, we compute the optical flow for the selected $R$ ROIs between the first frame $c_i^1$ and frame $c_i^s$ to obtain MDMO optical flow features $o_i^s=[o_i^{s,r}]_{r=1}^R\in\mathbb{R}^{R\times 2}$. The computation is similar to Eq. (\ref{of}).
Afterwards, we concatenate the optical flow $[o_i^1, o_i^2, \ldots, o_i^w]$ to construct multi-temporal-resolution optical flow features $o_i\in\mathbb{R}^{w\times R\times 2}$ for the clip $c_i$, where $o_i^1=\bm{0}$.
As a result of the data pre-processing, we concatenate the multi-temporal-resolution optical flow from all clips and obtain the compact sliding-window-based multi-temporal-resolution optical flow (CSW-MRO) features $O=(o_i)_{i=1}^N\in\mathbb{R}^{N\times w\times R\times 2}$ for the entire input video.

It is important to emphasize that the proposed CSW-MRO feature is not a simple reuse of existing techniques, but a task-driven redesign tailored for FES.
Existing methods adopt diverse strategies for optical flow computation. In many deep learning methods that compute optical flow for temporal modeling, it is often computed between adjacent frames, which provides robustness to head movement but may limit the magnification of motion information for detecting MEs. Moreover, the temporal range for optical flow modeling is either implicitly determined by the network architecture (e.g., the receptive fields of designed CNNs) or explicitly specified as a predefined hyper-parameter (e.g., in pretrained I3D-based methods during feature extraction), without explicit investigation into its optimality for distinguishing MaEs and MEs~\cite{pan2021spatio, wang2021mesnet, yu2021lssnet, leng2022abpn, yin2023aware, yu2023lgsnet, guo2023micro, li2024learning, yang2026mul}. In contrast, some recent traditional methods explore computing optical flow between a reference frame and subsequent frames over long temporal windows, which can enhance motion magnitude but may be more sensitive to head movements~\cite{yuhong2021research, yu2022facial, zhao2022rethinking, qin2023micro, yu2023efficient, yang2025ofct}.


We identify these limitations and combine the complementary advantages of both strategies by computing multi-temporal-resolution optical flow within compact sliding windows, which explicitly balances motion magnification and robustness to head movement. In addition, we further investigate the impact of window length and provide empirical insights into its role in distinguishing MaEs and MEs (see Section~\ref{temwindowsize}), thereby enabling accurate frame-level probability estimation for FES.


\subsection{SpotFormer for probability estimation}
The overview of the proposed SpotFormer is outlined in Fig.~\ref{model1}. After obtaining the optical flow features, SpotFormer calculates the frame-level probabilities required for expression spotting. These include the probabilities of occurrence for onset, apex, offset, expression, and neutral states for each frame in both ME and MaE spotting. SpotFormer consists of graph node embedding, temporal positional embedding, spatial facial embedding, and multi-scale spatio-temporal attention block. All these modules enable SpotFormer to comprehensively learn both spatial relations and temporal variations, facilitating accurate estimation of the probability map for every single frame. Subsequently, we will introduce each component in detail.

\subsubsection{Graph node embedding}
Given an input clip of optical flow features $o_i\in\mathbb{R}^{w\times R\times 2}$ for evaluating the frame $v_i$, we treat each graph node as a patch and use a trainable linear projection to embed each patch into $D$ dimensions.

\subsubsection{Temporal positional embedding}
For the temporal dimension, following \cite{dosovitskiy2021an}, we set tokens with learnable parameters, denoted as $\delta_{\mathrm{t}} \in \mathbb{R}^{w\times D}$, to retain temporal positional information. Note that all graph patches in a single frame share the same temporal positional embedding.

\subsubsection{Spatial facial embedding}
Unlike other graph-based tasks where there are predefined edge connections among graph nodes, facial expression analysis involves many possible edge connections among different facial muscles. 
To mitigate the impact of complex and diverse edge connections, SpotFormer constructs parameterized associations between nodes based on facial muscle characteristics, unlike traditional GCNs that rely on pre-defined edge connections. 
Inspired by \cite{zhang2021STST}, we introduce learnable facial part tokens to the graph nodes. Specifically, we set three tokens with learnable parameters $\delta_\mathrm{s} \in \mathbb{R}^{3\times D}$ for three asymmetric facial parts (left eyebrow, right eyebrow, and mouth) since MEs are asymmetric. Graph nodes belonging to the same facial part share the same facial part token. This method enables the model to understand the overall spatial facial structure without the requirement of exploring complex pre-defined edge connections.

\subsubsection{Spatio-temporal attention}
In the spatial graph attention layer, our objective is to model the spatial correlations among all graph nodes in each frame.
As shown in Fig.~\ref{model1} (c), the computation of each spatial graph attention layer can be described as follows:
\begin{equation}
\begin{aligned}
    &z^l=\mathrm{MHSA}(\mathrm{BN}(z^{l-1}))+z^{l-1},\\
    &z^{l+1}=\mathrm{MLP}(\mathrm{BN}(z^{l}))+z^{l},
\end{aligned}
\end{equation}
where $z^{l-1}$ denotes the output features of the last attention layer, $\mathrm{BN(\cdot)}$, $\mathrm{MHSA}(\cdot)$, and $\mathrm{MLP}(\cdot)$ denote batch normalization (BN) layers, multi-head self-attention (MHSA), and multilayer perceptron (MLP) blocks, respectively. The spatial attention layer allows the model to emphasize relevant spatial information and effectively capture the complex relationships among different facial parts.

In the temporal node attention layer, the computation process is similar, but we apply MHSA to each graph node across all frames within a compact sliding window, which enables the model to focus on temporal dependencies, facilitating the understanding of temporal variations within the sliding window.

\subsubsection{Multi-scale spatio-temporal feature learning}
Multi-scale learning has shown powerful performance in image processing \cite{lin2017feature, liu2021swin} and video understanding \cite{Fan_2021_ICCV, Shao_2023_ICCV}. It is also significant in facial expression analysis since it enables the model to extract both coarse and fine feature representations across different scales, enhancing its ability to capture comprehensive facial structures and temporal dynamics.

SpotFormer extracts multi-scale spatio-temporal features, enabling more comprehensive learning of motion patterns from low-level to high-level.
Specifically, to extract multi-scale temporal features, we use 1D CNNs with a stride of 2 for temporal downsampling, as proposed in ActionFormer \cite{zhang2022actionformer}. 

However, for spatial downsampling, applying pooling operations, which are generally used for downsampling images, to graph-structured data presents challenges because it is a type of non-Euclidean structured data. To address this issue, inspired by Xu et al. \cite{xu2021graph}, we introduce FLGP, specifically designed for extracting multi-scale facial graph features. In practice, we design three scales of facial structures. The designed scales and the scale change achieved through FLGP are illustrated in Fig.~\ref{model1} (b). During each FLGP operation, the facial graph is downsampled by aggregating features from multiple nodes using the max pooling operation.

\subsubsection{Model architecture}

SpotFormer simultaneously models spatio-temporal relations, as shown in Fig.~\ref{model1} (a). Specifically, in each spatio-temporal attention block, the model first employs a spatial graph attention layer to model intricate spatial relationships, followed by a temporal node attention layer that captures temporal variations among each graph node across frames. After each spatio-temporal attention block, spatial downsampling and temporal downsampling are performed to aggregate information to generate high-level features. Note that when the facial graph structure is downsampled to one node, subsequent temporal information learning relies solely on temporal node attention layers and temporal downsampling until the temporal dimension is reduced to 1.

\subsubsection{Frame-level probability estimation}
Finally, after reducing the spatial graph to a single node and temporal dimension to 1 through SpotFormer, the remaining single node comprehensively aggregates spatio-temporal information within a compact sliding window. We then use a fully-connected (FC) layer to generate the probability map $p_i=\{p_i^{\mathrm{apex}}$, $p_i^{\mathrm{exp}}$, $p_i^{\mathrm{onset}}$, $p_i^{\mathrm{offset}}$, $p_i^{\mathrm{nb}}$\} corresponding to the frame $v_i$. This map contains the probabilities of $v_i$ being an apex frame, expression frame, onset frame, offset frame, or non-boundary frame. Additionally, each component in $p_i$ incorporates two probabilities for ME spotting and MaE spotting, respectively. Specifically, $p_i^{\mathrm{apex}}=\{p_i^{\mathrm{me},\mathrm{apex}}, p_i^{\mathrm{mae},\mathrm{apex}}\}$, $p_i^{\mathrm{exp}}=\{p_i^{\mathrm{me},\mathrm{exp}}, p_i^{\mathrm{mae},\mathrm{exp}}\}$, $p_i^{\mathrm{onset}}=\{p_i^{\mathrm{me},\mathrm{onset}}, p_i^{\mathrm{mae},\mathrm{onset}}\}$, $p_i^{\mathrm{offset}}=\{p_i^{\mathrm{me},\mathrm{offset}}, p_i^{\mathrm{mae},\mathrm{offset}}\}$, $p_i^{\mathrm{nb}}=\{p_i^{\mathrm{me},\mathrm{nb}}, p_i^{\mathrm{mae},\mathrm{nb}}\}$. Therefore, the model outputs ten probabilities for each frame. The optimization tasks are divided into two groups for MaE spotting and ME spotting, respectively. Each group contains two binary classification tasks and a three-class classification task for different types of frames, following the optimization method outlined in \cite{leng2022abpn}. 
Specifically, the first binary classification task uses $p_i^{\mathrm{apex}}$ to determine whether the current frame is an apex frame. The second binary classification task uses $p_i^{\mathrm{exp}}$ to determine whether it is an expression or a neutral frame.
The three-class classification task uses $p_i^{\mathrm{onset}}$, $p_i^{\mathrm{offset}}$, and $p_i^{\mathrm{nb}}$ to classify the frame as onset, offset, or non-boundary.

Focal-loss \cite{lin2017focal} is employed to optimize our model, which can be expressed as:
\begin{equation}
    \mathcal{L}_{\mathrm{cls}} = -\sum_{i} y_i\alpha (1 - p_i)^\gamma \log(p_i),
\end{equation}
where $y_i$ is the ground-truth label, $\alpha$ and $\gamma$ are hyperparameters, respectively.

\subsection{Supervised contrastive learning}
We have been focusing on minimizing the divergence between predicted class probabilities and ground-truth class labels so far, potentially overlooking distributional differences among various classes. We notice that distinguishing certain MaEs and MEs near the boundary poses challenges in terms of duration and intensity. Specifically, the annotation of expressions follows the criterion that MaEs are longer than 0.5 seconds, while MEs are below 0.5 seconds. This criterion introduces difficulties in distinguishing expressions whose duration is close to 0.5 seconds. Similar challenges occur when distinguishing between ME frames and neutral frames. This is due to certain ground-truth MEs exhibiting very low intensity, which can be easily confused with noise in the optical flow features, leading to the misclassification of some neutral frames as ME frames.

To overcome this issue, we introduce supervised contrastive learning \cite{khosla2020supervised} to enhance our model's discriminative feature learning, aiming to improve its ability to recognize the classification boundaries between different types of expressions at the frame level. Specifically, we utilize the feature representation before the last classification FC layer as the feature representation for each frame. The supervised contrastive loss is then employed to minimize the distance between feature representations of the same expression class while simultaneously pushing apart feature representations of different expression classes. The expression type label $\widetilde{y}_i$ of the frame $v_i$ for the $i$-th sliding window $c_i$ is used as the supervision information for the supervised contrastive loss. This means that each frame is labeled as a MaE frame, ME frame, or neutral frame. Let $I$ denote a set of samples in a batch, and the loss function can be expressed as:
\begin{equation}
\begin{aligned}
\mathcal{L}_{\mathrm{con}}=\sum_{i\in I}\frac{-1}{|Q(i)|}\sum_{q\in Q(i)}\log\frac{\exp(z_i\cdot z_q/\tau)}{\sum_{e\in E(i)}\exp(z_i\cdot z_e/\tau)},
\label{supconloss}
\end{aligned}
\end{equation}
where $E(i) \coloneqq I\setminus i$, $Q(i) \coloneqq \{q\in E(i) \mid \widetilde{y}_q=\widetilde{y}_i\}$ represents the set of samples in the batch who has the same label with the $i$-th sample, $\tau \in \mathbb{R}^+$ is a scalar temperature parameter, and $z_i$ is the feature representation of $i$ which is extracted from the network.
The overall loss function for the optimization of our model can be formulated as follows:
\begin{equation}
    \mathcal{L} = \mathcal{L}_{\mathrm{cls}} + \lambda\mathcal{L}_{\mathrm{con}},
    \label{sumloss}
\end{equation}
where $\lambda$ is a weight parameter to balance between classification and supervised contrastive learning.

\begin{table*}[htbp]
\setlength\tabcolsep{4pt}
  \centering
  \caption{Comparison with the state-of-the-art methods on CAS(ME)$^2$ and SAMM-LV in terms of F1-score.}
    \begin{tabular*}{0.75\hsize}{@{}@{\extracolsep{\fill}}llcccccc@{}}
    \hline
    \multicolumn{2}{c}{Methods} & \multicolumn{3}{c}{SAMM-LV} & \multicolumn{3}{c}{CAS(ME)$^2$} \\
    &&MaE&ME&Overall&MaE&ME&Overall \\
    \hline
    Traditional methods&MDMD\cite{he2020spotting}&0.0629 &0.0364&0.0445&0.1196&0.0082&0.0376\\
    &Optical Strain\cite{gan2020optical}&-&-&-&0.1436&0.0098&0.0448\\
    &Zhang et al.\cite{zhang2020spatio}&0.0725&0.1331&0.0999&0.2131&0.0547&0.1403\\
    &He \cite{yuhong2021research}&0.4149&0.2162&0.3638&0.3782&0.1965&0.3436\\
    &Zhao et al.\cite{zhao2022rethinking}&-&-&0.3863&-&-&0.4030\\
    &Wang et al.\cite{wangunique}&0.3724&0.2866&0.3419&0.5061&0.2614&0.4558\\

    \hline
    Deep-learning methods&Verburg\cite{verburg2019micro}&-&0.0821&-&-&-&-\\
    &LBCNN\cite{pan2020local}&-&-&0.0813&-&-&0.0595\\
    &MESNet\cite{wang2021mesnet}&-&0.0880&-&-&0.0360&-\\
    &SOFTNet\cite{liong2021shallow}&0.2169&0.1520&0.1881&0.2410&0.1173&0.2022\\
    &3D-CNN\cite{yap20223d}&0.1595&0.0466&0.1084&0.2145&0.0714&0.1675\\
    &Concat-CNN\cite{yang2021facial}&0.3553&0.1155&0.2736&0.2505&0.0153&0.2019\\
    &LSSNet\cite{yu2021lssnet}&0.2810&0.1310&0.2380&0.3770&0.0420&0.3250\\
    &MTSN\cite{liong2022mtsn}&0.3459&0.0878&0.2867&0.4104&0.0808&0.3620\\
    &ABPN\cite{leng2022abpn}&0.3349&0.1689&0.2908&0.3357&0.1590&0.3117\\
    &AUW-GCN\cite{yin2023aware}&0.4293&0.1984&0.3728&0.4235&0.1538&0.3834\\
    &LGSNet\cite{yu2023lgsnet}&-&-&0.3880&-&-&0.4360\\
    &MULT\cite{GUO2023146}&-&0.2770&-&-&0.1373&-\\
    &\textbf{Ours}&\textbf{0.4447}&\textbf{0.4281}&\textbf{0.4401}&\textbf{0.5061}&\textbf{0.2817}&\textbf{0.4841}\\
    \hline
    \end{tabular*}%
  \label{results1}%
\end{table*}%

\subsection{Post-processing}
Once the series of output probabilities $P=(p_i)_{i=1}^{N}$ for the entire video $V$ are obtained, we perform MaE spotting and ME spotting independently, following the methodology outlined in \cite{leng2022abpn}. We will elaborate on the ME spotting process, with the MaE spotting process following a similar method. Initially, we identify all potential ME apex frames $U^{\mathrm{me},\mathrm{apex}}$ based on the criterion $p_l^{\mathrm{me}, \mathrm{apex}} > \theta_{\mathrm{apex}}$, where $\theta_{\mathrm{apex}}$ is a threshold. For each spotted apex frame $u^{\mathrm{me}, \mathrm{apex}}_l\in U^{\mathrm{me},\mathrm{apex}}$ with the frame index $l$, we select the onset frame with the highest onset probability from the left side of the apex frame within the range of $[l-\frac{k^{\mathrm{me}}}{2}, l-\frac{j^{\mathrm{me}}}{2}]$. Similarly, we select the offset frame with the highest offset probability from the right side of the apex frame within the range of $[l+\frac{j^{\mathrm{me}}}{2}, l+\frac{k^{\mathrm{me}}}{2}]$, where $k^{\mathrm{me}}$ and $j^{\mathrm{me}}$ define the maximum and minimum temporal ranges for searching boundaries around the apex frame, respectively. As a result, a ME proposal $\phi_l$ is obtained, including the onset frame, offset frame, and expression type. Subsequently, we assign a score $s_l=p_b^{\mathrm{me},\mathrm{onset}}\times p_l^{\mathrm{me},\mathrm{apex}} \times p_d^{\mathrm{me},\mathrm{offset}}$ to $\phi_l$, where $b$ and $d$ denote the frame indices of the onset frame and offset frame selected by the aforementioned rule, respectively.

After obtaining all possible expression proposals, we apply non-maximum suppression to eliminate redundant proposals. Specifically, if the overlap rate between two proposals exceeds $\theta_{\mathrm{overlap}}$, we compare the assigned scores and discard the proposal with the lower score, thereby obtaining the final spotting results.

\section{Experiments}
In this section, we first introduce the experimental setup (Section \ref{esetup})
used in the paper. We then compare our method with state-of-the-art methods to demonstrate the effectiveness of SpotFormer (Section \ref{comparesota}). Next, we thoroughly evaluate our proposed modules in SpotFormer in the ablation study (Section \ref{ablation}). We then explore several model variants and compare them with SpotFormer to investigate its effectiveness.
(Section \ref{modelvariations}). Finally, we present the detailed expression spotting results and a discussion of the results (Section \ref{secd}).

\subsection{Experimental setup}
\label{esetup}
\textbf{Datasets.}
We conduct experiments on three datasets: SAMM-LV \cite{yap2020samm}, CAS(ME)$^2$ \cite{qu2017cas}, and CAS(ME)$^\mathrm{3}$~\cite{9774929}. The SAMM-LV dataset includes 147 raw long videos with 343 MaE clips and 159 ME clips. The CAS(ME)$^2$ dataset includes 87 raw long videos with 300 MaE clips and 57 ME clips. The large-scale dataset CAS(ME)$^3$ includes
956 raw long videos with 3490 MaE clips and 1109 ME clips. Since the frame rate of the SAMM-LV dataset is 200fps while the frame rate of the CAS(ME)$^2$ and CAS(ME)$^3$ datasets is 30fps, we subsample every 7th frame from the SAMM dataset to align the frame rates of all three datasets.

\textbf{Evaluation metric.}
Following the protocol of MEGC2021~\cite{li2021fme}, we evaluate our model on each dataset separately and employ a leave-one-subject-out cross-validation strategy in our experiments. The true positive (TP) is defined based on the Intersection over Union (IoU) between the proposal and the ground-truth expression clip. Specifically, given a ground-truth expression clip and its expression type, we compare it with all expression proposals that have the same estimated expression type. An expression proposal $W_{\mathrm{Proposal}}$, is considered a TP when it satisfies the following condition:
\begin{equation}
\begin{aligned}
    \frac{W_{\mathrm{Proposal}}\cap W_{\mathrm{GroundTruth}}}{W_{\mathrm{Proposal}}\cup W_{\mathrm{GroundTruth}}}\geq\theta_{\mathrm{IoU}},
\end{aligned}
\end{equation}
where $\theta_{\mathrm{IoU}}$ is the IoU threshold, set to 0.5, and $W_{\mathrm{GroundTruth}}$ represents the ground-truth expression proposal (from the onset frame to the offset frame). Otherwise, the proposed expression proposal is considered a false positive (FP). A ground-truth expression clip is counted as a false negative (FN) when it does not match any expression proposal. Note that each ground-truth expression clip corresponds to at most one TP. We calculate the precision rate, recall rate, and F1 score to evaluate the performance of our model and compare it with other methods.

\textbf{Training details.} We use the AdamW optimizer \cite{loshchilov2017decoupled} to optimize our model, setting the learning rate to 0.0002, $\beta_{1}$ to 0.7, and $\beta_{2}$ to 0.9. $\theta_{\mathrm{apex}}$ is set to 0.5. The temperature parameter $\tau$ in Eq. (\ref{supconloss}) is set to 0.5. We train our model for 100 epochs with a batch size of 512. The facial images input for the CSW-MRO feature extraction are resized to $(320, 320)$. 
The maximum temporal ranges for searching boundaries around the apex frame, $k^{\mathrm{mae}}$ and $k^{\mathrm{me}}$, are set to 60 and 16, respectively, while the minimum temporal ranges, $j^{\mathrm{mae}}$ and $j^{\mathrm{me}}$, are set to 16 and 4. The hidden dimension and the number of heads in the MHSA mechanism is set to 256 and 4, respectively.

\begin{figure*}[tbp]
\centering
\includegraphics[width=0.95\textwidth]{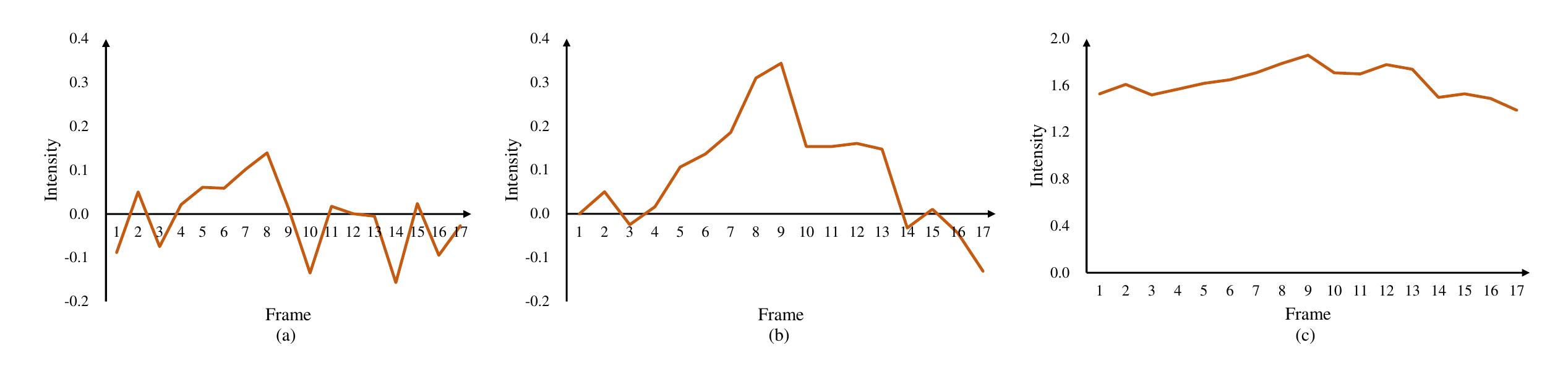}
\caption{Some visualization optical flow of certain micro-expression frames computed by three strategies. The data comes from the vertical component of optical flow computed at the left mouth corner when subject 11 from the SAMM-LV dataset is performing a micro-expression. (a) optical flow computed between adjacent frames; (b) optical flow computed using the proposed CSW-MRO; (c) optical flow computed with a large sliding window strategy.}
\label{fig9}
\end{figure*}

\subsection{Comparison with state-of-the-art methods}
\label{comparesota}
We first compare our method with the state-of-the-art (SOTA) methods on the SAMM-LV and CAS(ME)$^2$ datasets, and the results are shown in Table \ref{results1}. We report the F1-score for MaE spotting, ME spotting, and overall performance. For the overall performance, our method achieves F1-scores of 0.4401 on the SAMM-LV dataset and 0.4841 on the CAS(ME)$^2$ dataset, which outperforms other SOTA methods by 13.4\% and 6.2\%, respectively. For the MaE spotting, our method achieves an improvement of 3.6\% on the SAMM-LV dataset compared to other methods. It is important to emphasize our method's remarkable effectiveness in ME spotting. The results demonstrate a substantial enhancement with a 49.4\% improvement on the SAMM-LV dataset and a 7.8\% improvement on the CAS(ME)$^2$ dataset compared to other SOTA methods.

The remarkable improvements in ME spotting are mainly attributed to our proposed CSW-MRO feature. Previous methods generally extract optical flow between adjacent frames and employ either a pretrained I3D model or simple neural networks to model these coarse motion patterns. In contrast, our proposed CSW-MRO feature effectively extracts subtle motions while avoiding the optical flow being dominated by head movements. Moreover, the carefully designed SpotFormer can comprehensively analyze the latent motion patterns within the CSW-MRO feature. In addition, the proposed supervised contrastive learning further enhances discriminative feature learning between different expression types, leading to improved model performance. 

We further validate our method on the CAS(ME)$^3$ dataset \cite{9774929} and the results are shown in Table \ref{casme3}. Following LGSNet \cite{yu2023lgsnet}, we pre-process the dataset by considering only ground-truth MEs with a duration of less than 15 frames and ground-truth MaEs with a duration between 16 and 120 frames for training and validation, ensuring a fair comparison. This protocol is consistent with those adopted by the SAMM and CAS(ME)$^2$ datasets and also aligns with the general definition that MEs last less than 0.5 seconds, whereas MaEs typically last from 0.5 to 4.0 seconds. The data pre-processing results in 2231 ground-truth MaEs and 285 ground-truth MEs. The results demonstrate that our method outperforms LGSNet \cite{yu2023lgsnet}, especially in ME spotting, highlighting the strong generalization ability of our method.

\begin{table}[tbp]
\setlength\tabcolsep{4pt}
  \centering
  \caption{Comparison with the state-of-the-art method on CAS(ME)$^3$ in terms of F1-score.}
    \begin{tabular}{cccc} 
    \hline
    Methods&\multicolumn{3}{c}{CAS(ME)$^3$}\\
    &MaE&ME&Overall\\
    \hline
    LGSNet \cite{yu2023lgsnet}&-&0.0990&0.2350\\
    \textbf{Ours}&\textbf{0.2664}&\textbf{0.2037}&\textbf{0.2559}\\
    \hline
    \end{tabular}%
  \label{casme3}%
\end{table}%

\subsection{Ablation studies}
\label{ablation}
\textbf{Proposed CSW-MRO and FLGP.} We conduct ablation studies to evaluate our proposed modules and Table \ref{modules} shows the experimental results. The acronyms \textbf{CSW-MRO} and \textbf{FLGP} denote specific modules in our model. Specifically, \textbf{CSW-MRO} involves calculating compact sliding-window-based 
multi-temporal-resolution optical flow features instead of optical flow between adjacent frames, while \textbf{FLGP} integrates facial local graph pooling operation into SpotFormer for multi-scale spatial feature learning instead of global average pooling over facial graphs. The results show the effectiveness of each proposed module. Notably, the introduction of the CSW-MRO enhances the extracted motion features, particularly in revealing subtle motions that exist in MEs. 
This enhancement leads to a significant overall performance improvement of 17.0\%/17.7\% on the SAMM-LV and CAS(ME)$^2$ datasets. Especially in ME spotting, it has led to an improvement of 69.2\%/200.1\% on the SAMM-LV and CAS(ME)$^2$ datasets.
The introduction of the FLGP operation for multi-scale spatial feature learning results in a further improvement of 3.0\%/3.4\% on the SAMM-LV and CAS(ME)$^2$ datasets compared to using global average pooling. This demonstrates the superior representational capabilities of our proposed model.

\begin{table}[tbp]
\setlength\tabcolsep{3pt}
  \centering
  \caption{Results of the ablation study on the effectiveness of the proposed modules.}
    \begin{tabular}{cccccccc} 
    \hline
    & &\multicolumn{3}{c}{SAMM-LV}&\multicolumn{3}{c}{CAS(ME)$^2$}\\
    CSW-MRO&FLGP&MaE&ME&Overall&MaE&ME&Overall\\
    \hline
    \ding{55}&\ding{55}&0.4062&0.2424&0.3687&0.4441&0.0625&0.4105\\
    \ding{55}&\Checkmark&0.4196&0.2530&0.3760&0.4431&0.0936&0.4113\\
    \Checkmark&\ding{55}&0.4418&0.3889&0.4274&0.4907&0.2462&0.4684\\
    \Checkmark&\Checkmark&0.4447&0.4281&0.4401&0.5061&0.2817&0.4841\\
    \hline
    \end{tabular}%
  \label{modules}%
\end{table}%

Fig. \ref{fig9} shows the qualitative comparison between different optical flow extraction strategies. When computing optical flow between adjacent frames, the motions present in MEs are so subtle that the noise in the optical flow could overshadow these subtle movements, impacting the quality of the data and making it difficult to reveal these subtle expressions.
On the other hand, employing a large sliding window strategy for optical flow computation introduces substantial interference from irrelevant motions such as head movements, posing challenges in accurately detecting facial expressions. On the contrary, the utilization of the CSW-MRO helps alleviate these problems. These results further confirm that the proposed CSW-MRO feature is not a simple adoption of existing optical flow extraction strategies, but a task-specific design for FES. It achieves a balance between amplifying subtle facial muscle motions and mitigating the impact of head movements. 
This design is particularly suited for frame-level probability-estimation-based deep learning frameworks, as such models do not require large temporal receptive fields to perceive complete MaEs. Instead, compact windows are sufficient to capture complete MEs while distinguishing between MEs and MaEs. Further discussion is provided in Section~\ref{temwindowsize}.

\begin{figure}[tbp]
\centering
\includegraphics[width=0.5\textwidth]{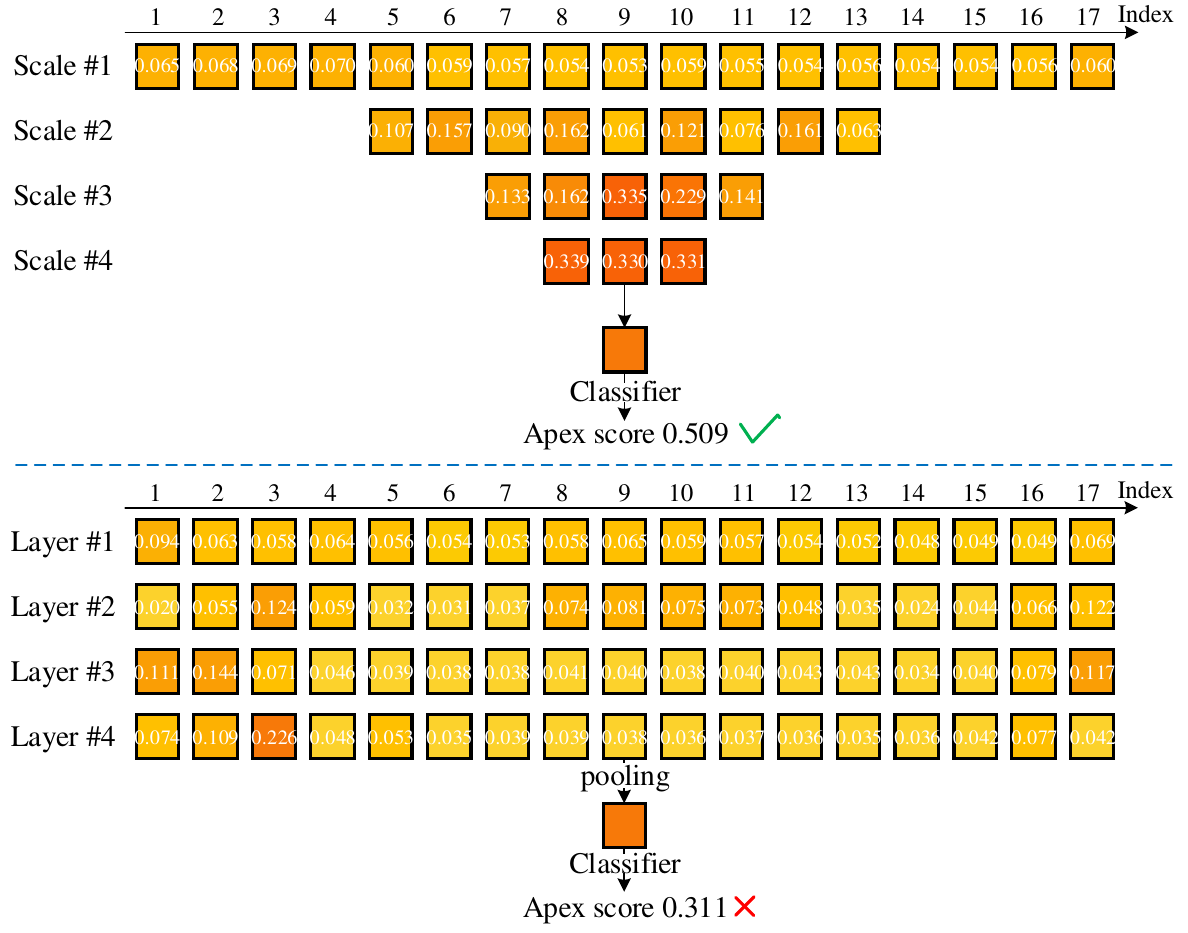}
\caption{Comparison of temporal attention maps with and without multi-scale temporal modeling. The predicted frame corresponds to a ground-truth ME apex frame.}
\label{fig:multiscale}
\end{figure}

\begin{figure*}[tbp]
\centering
\includegraphics[width=1.0\textwidth]{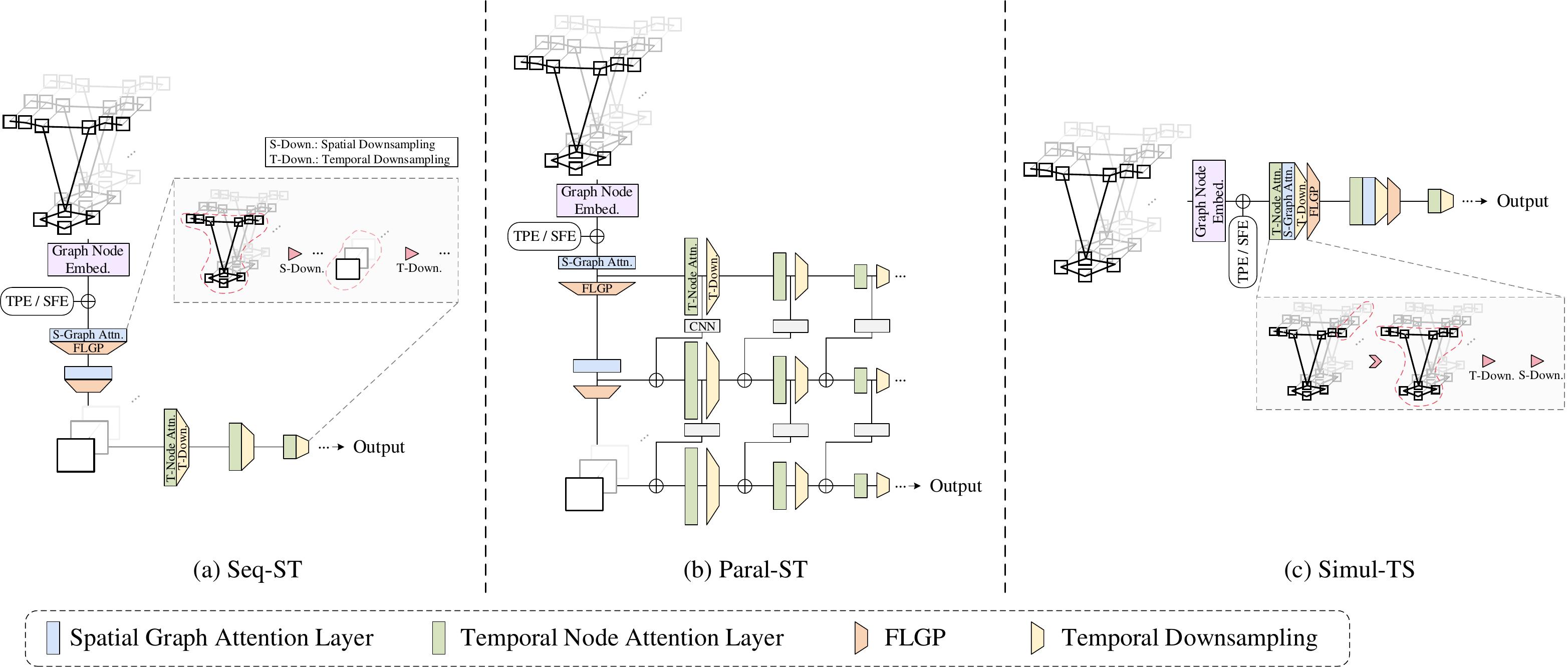}
\caption{Overview of the variants of the proposed model architectures. (a) \textit{Seq-ST} first employ spatial graph attention layers and FLGP operations to encode the graph of each frame to a single node, and then models temporal variations across frames; (b) \textit{Paral-ST} differs from \textit{Seq-ST} by modeling temporal variations at each graph scale and performing multi-scale feature fusion; (c) \textit{Simul-TS} differs from SpotFormer by modeling temporal variations first and subsequently learning spatial relationships within each spatio-temporal attention block.}
\label{3models}
\end{figure*}

\noindent\textbf{Multi-scale temporal modeling.}
Although 1D CNN-based temporal downsampling has been widely adopted for multi-scale temporal feature extraction~\cite{zhang2022actionformer}, we explicitly evaluate its necessity in our framework. Specifically, we remove progressive temporal downsampling by setting the stride of all temporal CNN layers to 1, while keeping the attention architecture unchanged, and apply a single temporal pooling operation only after all attention layers. As shown in Table~\ref{tempmodel}, the performance of MaE spotting remains relatively stable under this single-scale setting due to the longer duration of MaEs. In contrast, ME spotting shows a substantial performance drop. This demonstrates that relying on a fixed temporal resolution is insufficient for effectively capturing subtle and short motion transients of MEs. Without multi-scale temporal modeling, these transient patterns become less salient during long-range temporal aggregation, as they may occupy only a small fraction of the input sequence. In contrast, progressive temporal downsampling facilitates hierarchical temporal abstraction, which better integrates discriminative cues for expressions of varying durations.

An example is shown in Fig.~\ref{fig:multiscale}, where, without multi-scale temporal modeling, attention remains diffused across the full temporal resolution, while progressive temporal downsampling gradually concentrates attention on coarse temporal regions corresponding to the ME, thereby enabling the detection of the ME apex frame. Both quantitative and qualitative results indicate that proper multi-scale temporal modeling is critical for accurate ME spotting.

\begin{table}[tbp]
\setlength\tabcolsep{1pt}
  \centering
  \caption{Results of the ablation study on the multi-scale temporal modeling.}
    \begin{tabular}{ccccccc} 
    \hline
    &\multicolumn{3}{c}{SAMM-LV}&\multicolumn{3}{c}{CAS(ME)$^2$}\\
    Temporal downsampling&MaE&ME&Overall&MaE&ME&Overall\\
    \hline
    Average pooling&0.4286&0.3755&0.4139&0.4917&0.1515&0.4609\\
    Max pooling&0.4349&0.3113&0.4009&\textbf{0.5110}&0.0345&0.4713\\
    Multi-scale (Ours)&\textbf{0.4447}&\textbf{0.4281}&\textbf{0.4401}&0.5061&\textbf{0.2817}&\textbf{0.4841}\\
    \hline
    \end{tabular}%
  \label{tempmodel}%
\end{table}%

\subsection{Model variations}
\label{modelvariations}
\subsubsection{Multiple model architectures} 
To verify the effectiveness of SpotFormer, we explored multiple model architectures that consider multi-scale spatio-temporal feature learning differently as variants of SpotFormer and then conducted a comprehensive comparison among them. Here, the original architecture of SpotFormer is called \textit{Simul-ST} for clarity since it simultaneously models spatio-temporal relations and enables the comprehensive capture of both spatial and temporal information. The architectures of variants are described as follows:

\textbf{Seq-ST.} \textit{Seq-ST} sequentially learns spatial features first, followed by temporal features. As shown in Fig. \ref{3models} (a), we first employ two spatial graph attention layers, each followed by an FLGP operation for spatial downsampling. After reducing the facial graph structure to one node, we start adding temporal node attention layers, where each is followed by a temporal downsampling layer until temporal dimension is downsampled to 1. \textit{Seq-ST} is similar to a general TAL architecture, which encodes spatial information into a feature vector and then learns temporal information to detect potential action instances. In both TAL and FES, temporal information is vital for detecting action instances; such a model can initially focus on spatial details, capturing fine-grained information, and then capturing temporal dynamics.



\textbf{Paral-ST.} 
Considering that \textit{Seq-ST} lacks the consideration of temporal information in lower graph scales, \textit{Paral-ST} parallelizes temporal learning at each spatial scale and fuses them. Specifically, as described in Fig. \ref{3models} (b), we perform temporal node attention and temporal downsampling at each facial graph scale. 
At each temporal scale, we fuse temporal features from different graph scales by incorporating the output of the temporal node attention layer into the higher graph scale. The incorporation is achieved through a CNN layer followed by an element-wise addition operation.
In \textit{Paral-ST}, the independent extraction of temporal dynamics for each spatial graph scale and the multi-scale feature fusion ensure a comprehensive understanding of temporal dynamics.

\begin{table*}[htbp]
\setlength\tabcolsep{4pt}
  \centering
  \caption{Comparison of multiple model architectures on SAMM-LV and CAS(ME)$^2$.}
    \begin{tabular*}{0.75\hsize}{@{}@{\extracolsep{\fill}}lcccccccc@{}}
    \hline
    &\multicolumn{3}{c}{SAMM-LV}&\multicolumn{3}{c}{CAS(ME)$^2$}\\
    Model&MaE&ME&Overall&MaE&ME&Overall&Overall\\
    \hline
    SpotFormer (Simul-ST)&0.4447&\textbf{0.4281}&\textbf{0.4401}&0.5061&\textbf{0.2817}&0.4841&\textbf{0.4584}\\
    Seq-ST&\textbf{0.4680}&0.3630&0.4367&0.4922&0.2133&0.4630&0.4478\\
    Paral-ST&0.4235&0.3919&0.4139&\textbf{0.5368}&0.2353&\textbf{0.5054}&0.4506\\
    Paral-ST (no addi. para.)&0.4163&0.3378&0.3964&0.5170&0.0984&0.4794&0.4324\\
    Simul-TS&0.4459&0.3959&0.4321&0.4992&0.2121&0.4733&0.4489\\
    \hline
    \end{tabular*}%
  \label{models}%
\end{table*}%

\textbf{Simul-TS.} Would it be better to prioritize temporal node attention first? With such consideration, \textit{Simul-TS} begins by focusing on learning temporal dynamics and gradually introduces spatial feature learning. As described in Fig. \ref{3models} (c), in each spatio-temporal attention block, we first perform temporal node attention, followed by spatial graph attention. Subsequently, we perform temporal downsampling and spatial downsampling to generate high-level features. By learning spatial information later in the process, \textit{Simul-TS} treats spatial information as more important.

We validate and compare our proposed multiple multi-scale spatio-temporal model architectures on the SAMM and CAS(ME)$^2$ datasets, and the experimental results are shown in Table \ref{models}. With the exception of \textit{Paral-ST}, our models use the same number of parameters for a fair comparison. 

The results indicate that SpotFormer (\textit{Simul-ST}) achieves the best overall performance, as it learns spatial and temporal information simultaneously. \textit{Seq-ST} is similar to a TAL model, encoding spatial information into a feature vector and then learning temporal information. The comparison results show that simultaneously learning spatio-temporal information better models the spatio-temporal relationships, leading to improved performance.
\textit{Paral-ST (no more parameter)} considers multi-scale feature fusion without adding additional parameters, we implement it by removing additional CNN layers and using weight-sharing temporal node attention layers at each temporal scale. However, the performance of \textit{Paral-ST (no more parameter)} is worse, especially in ME spotting. This is because redundant information is fed into the high-scale branch without being learned properly, thus harmful for expression spotting. \textit{Paral-ST}, with added parameters for processing the lower-scale features, increases the model's ability to handle multi-scale features. The results of \textit{Simul-TS} indicate that modeling spatial correlations first is better than modeling temporal variations first. This is because, unlike facial expression recognition, where we can first learn temporal information and then learn spatial information (AUs combination) to recognize the exact emotion type, facial expression spotting focuses on detecting the existing facial expressions, making temporal information more crucial.


\begin{table}[tbp]
\setlength\tabcolsep{4pt}
  \centering
  \caption{Comparison of different temporal window sizes (receptive fields) on SAMM-LV.}
    \begin{tabular}{cccc}
    \hline
    Window Size / & \multirow{2}{*}{MaE}&\multirow{2}{*}{ME}&\multirow{2}{*}{Overall} \\
      Receptive Field\\
    \hline
    9&0.4278&0.2419&0.3794\\
    11&0.4272&0.2709&0.3874\\
    15&0.4254&0.3883&0.4151\\
    \textbf{17}&0.4447&\textbf{0.4281}&\textbf{0.4401}\\
    19&0.4397&0.3922&0.4269\\
    21&\textbf{0.4557}&0.3851&0.4350\\
    25&0.4421&0.3408&0.4127\\
    \hline
    \end{tabular}%
  \label{results2}%
\end{table}%

\subsubsection{Temporal window size}
\label{temwindowsize}
We further explore how much temporal information is needed to accurately evaluate one frame. Specifically, we compare different temporal window sizes (i.e., temporal receptive field) for the CSW-MRO on the SAMM-LV dataset, and the experimental results are presented in Table \ref{results2}. The results indicate that the optimal performance is obtained when the temporal window size is set to 17, which demonstrates the effectiveness of our core idea: \textbf{perceiving complete MEs and distinguishing between MaEs and MEs}. Specifically, the temporal boundary for distinguishing MaEs and MEs is 0.5 seconds, which corresponds to 15 frames when the frame rate is set to 30 fps. A receptive field of 17 frames proves sufficient to perceive a complete ME and amplify the motion information that exists in MEs. When the model can accurately spot MEs, windows that exhibit greater intensity of motion, or windows that exhibit a continuous expression and are hard to perceive a trend of this expression occurring or ceasing within a short period, are more likely to contain MaEs.

\begin{table}[tbp]
\setlength\tabcolsep{4pt}
  \centering
  \caption{Results of the ablation study on supervised contrastive learning and the choice of hyperparameter.}
    \begin{tabular}{ccccccc} 
    \hline
    &\multicolumn{3}{c}{SAMM-LV}&\multicolumn{3}{c}{CAS(ME)$^2$}\\
    $\lambda$&MaE&ME&Overall&MaE&ME&Overall\\
    \hline
    0.0&0.4320&0.3840&0.4153&0.4495&0.2632&0.4164\\
    0.001&\textbf{0.4514}&0.3704&0.4240&0.4934&0.2716&0.4698\\
    0.003&0.4429&0.3873&0.4261&0.4947&0.2500&0.4704\\
    \textbf{0.005}&0.4447&\textbf{0.4281}&\textbf{0.4401}&\textbf{0.5061}&\textbf{0.2817}&\textbf{0.4841}\\
    0.008&0.4501&0.3881&0.4330&0.4954&0.2029&0.4675\\
    0.01&0.4509&0.3505&0.4230&0.5057&0.1875&0.4758\\
    0.03&0.4345&0.2621&0.3961&0.5024&0.0678&0.4651\\
    0.05&0.4363&0.2723&0.3983&0.4977&0.0345&0.4596\\
    \hline
    \end{tabular}%
  \label{results3}%
\end{table}%

\begin{figure*}[tbp]
\centering
\includegraphics[width=1.0\textwidth]{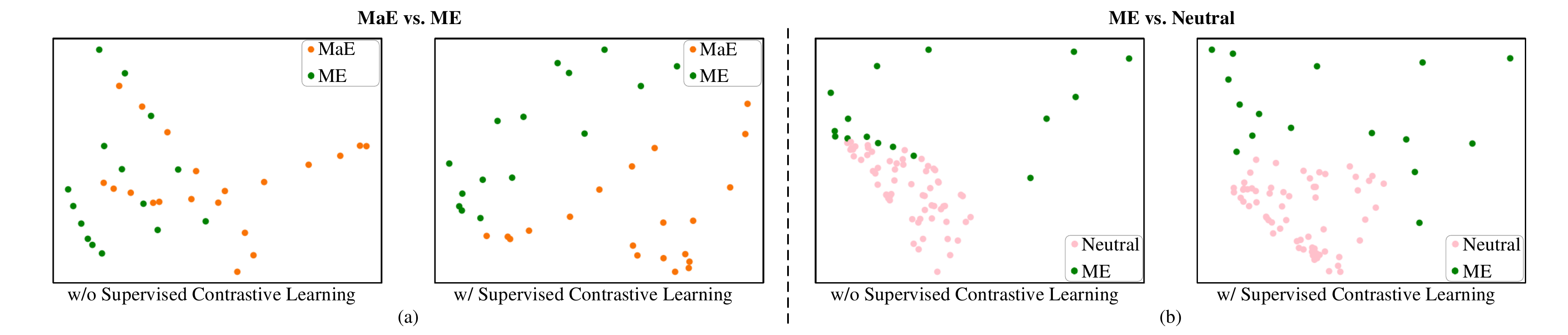}
\caption{Visualization analysis of supervised contrastive learning.  
(a) PCA distributions of certain MaE and ME frames: left, without supervised contrastive learning; right, with supervised contrastive learning.  
(b) PCA distributions of certain ME and neutral frames: left, without supervised contrastive learning; right, with supervised contrastive learning.
}
\label{fig6}
\end{figure*}

\begin{figure*}[tbp]
\centering
\includegraphics[width=0.9\textwidth]{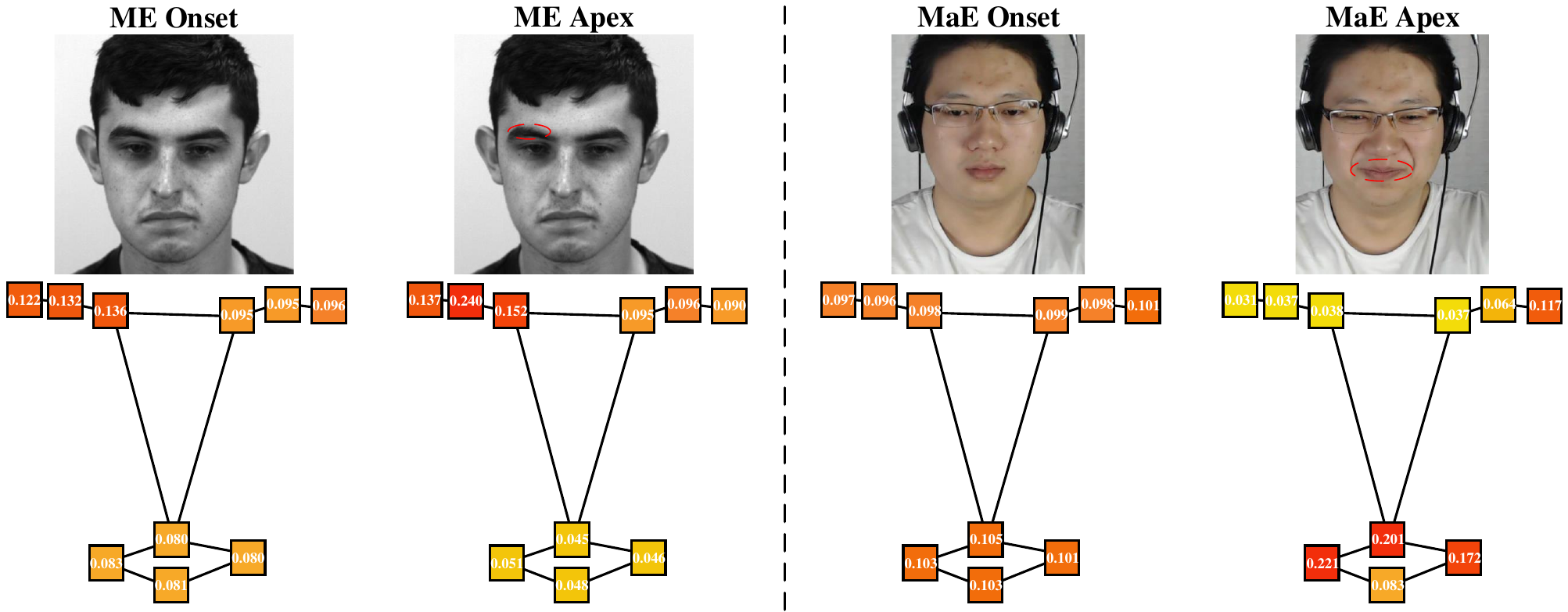}
\caption{Qualitative analysis of SpotFormer. The highlighted ROIs indicate the facial regions most relevant to expression dynamics and most influential to the model’s predictions.}
\label{roianalysis}
\end{figure*}

Based on this idea, the model can capture the temporal variations necessary to distinguish between general MaEs and MEs. A smaller window size cannot provide sufficient temporal information, while an increase in the window size results in more frames, potentially leading to information redundancy and increased ambiguity in distinguishing between MaEs and MEs. Moreover, larger temporal window sizes may cause the optical flow to be severely dominated by head movements, negatively impacting overall performance.

\subsubsection{Supervised contrastive loss and hyperparameter $\lambda$} Table \ref{results3} shows the experimental results of introducing supervised contrastive learning into our model and the impact of the hyperparameter $\lambda$ in Eq. (\ref{sumloss}), which is set to balance classification and supervised contrastive learning.
We observe that the introduction of the supervised contrastive loss enables our model to better recognize the classification boundaries between different types of expressions, resulting in an improvement of 6.0\%/16.3\% on the SAMM-LV and CAS(ME)$^2$ datasets. The optimal value for $\lambda$ is 0.005, increasing $\lambda$ beyond this value starts to impact standard classification, leading to a decrease in spotting accuracy. 

Fig.~\ref{fig6} shows the qualitative comparison between scenarios with and without supervised contrastive learning. In the inference phase, we randomly sampled frames and applied principal component analysis (PCA) \cite{bro2014principal} to analyze the distribution of various expression classes. Subsequently, we marked each frame with its corresponding ground-truth expression label. In Fig.~\ref{fig6} (a), we examine the PCA distribution of specific MaE frames and ME frames with and without supervised contrastive learning. Moving to Fig. \ref{fig6}~(b), we compare the PCA distribution of specific neutral frames and ME frames with and without supervised contrastive learning. 
The results indicate that, in the absence of supervised contrastive learning, our model may face challenges distinguishing certain MaE frames from ME frames due to subtle differences. This leads to mixed distributions and misclassification. 
Additionally, neutral frames may be mistakenly classified as ME frames, as the noise existing in the optical flow introduces slight fluctuations that the model may treat as ME motion information, resulting in incorrect ME proposal detection. With the introduction of supervised contrastive learning into our model, the domain discrepancy increases, thus significantly improving the accuracy of expression spotting.


\begin{figure*}[tbp]
\centering
\includegraphics[width=0.95\textwidth]{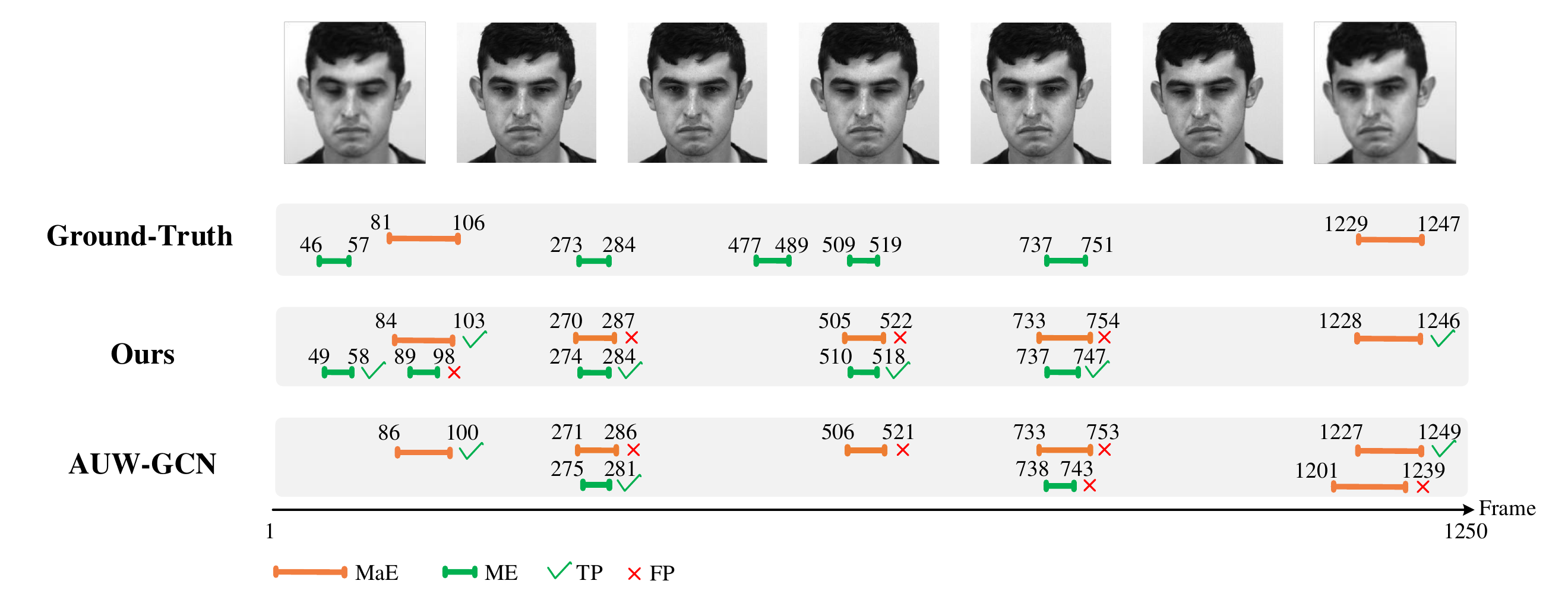}
\caption{Qualitative results of our method on SAMM-LV. The example video is ``$006\_1$", whose frame rate has been downsampled to 30fps.}
\label{videoresult}
\end{figure*}

\begin{table*}[htbp]
\setlength\tabcolsep{4pt}
  \centering
  \caption{Detailed spotting results of the proposed method on SAMM-LV and CAS(ME)$^2$.}
    \begin{tabular*}{0.75\hsize}{@{}@{\extracolsep{\fill}}llcccccc@{}}
    \hline
    Dataset & \multicolumn{3}{c}{SAMM-LV}&\multicolumn{3}{c}{CAS(ME)$^2$} \\
      Expression & MaE&ME&Overall&MaE&ME&Overall\\
    \hline
    Total&343&159&502&300&57&357\\
    TP&163&61&224&165&10&175\\
    FP&227&65&292&187&4&191\\
    FN&180&98&278&135&47&182\\
    Precision&0.4179&0.4841&0.4341&0.4688&0.7143&0.4781\\
    Recall&0.4752&0.3836&0.4462&0.5500&0.1754&0.4902\\
    F1-Score&0.4447&0.4281&0.4401&0.5061&0.2817&0.4841\\
    \hline
    \end{tabular*}%
  \label{results4}%
\end{table*}%

\subsection{Qualitative analysis}
To better understand the interpretability of SpotFormer, we visualize the ROI-level spatial attention maps of two representative samples, which indicate how strongly each facial region contributes to the model’s prediction at each time point. Each sample includes the attention maps of the onset and apex frames, together with their corresponding raw images. As shown in Fig.~\ref{roianalysis}, the highlighted ROIs correspond well to the facial areas involved in expression-related muscle activation. These visualizations demonstrate that SpotFormer can effectively localize the facial regions most relevant to expression dynamics, confirming its effectiveness in FES. The results further highlight the advantage of ROI-based modeling over methods that directly compress spatial information into a single feature vector, which may lose fine-grained local cues and become easily influenced by irrelevant motion noise.

\subsection{Detailed results}
\label{secd}
Table \ref{results4} shows the detailed results on the SAMM-LV and CAS(ME)$^2$ datasets. Our method achieves an overall precision rate of 0.4341/0.4781 and an overall recall rate of 0.4462/0.4902 on SAMM and CAS(ME)$^2$, respectively. Specifically, our method achieves a recall rate of approximately 0.5 in MaE spotting and 0.4 in ME spotting. The exception is for ME spotting on the CAS(ME)$^2$ dataset. The results indicate that it achieves a high precision rate while a low recall rate. The primary reason is that the scarcity and imbalance of ME data in the CAS(ME)$^2$ dataset restrict the model’s ability to learn sufficiently discriminative representations for subtle ME patterns. CAS(ME)$^2$ contains only 57 ME clips, and some samples involve only eye blinking—a motion generally considered irrelevant for FES. Due to the weak ME representations learned from such limited data, several true ME apex frames receive relatively low apex scores. Since our post-processing requires the apex probability to exceed a strict, dataset-independent threshold (0.5), these low-scoring true apex frames are filtered out, even though their scores remain higher than those of typical neutral frames. This results in missed detections and thus lower recall. Reducing the threshold specifically for ME spotting on CAS(ME)$^2$ increases TPs but also introduces more FPs, indicating that the core issue lies in insufficient ME representation learning rather than threshold design.

We also found that, on both datasets, the precision is lower than the recall in MaE spotting, while the opposite is observed in ME spotting. This is mainly due to data imbalance and the specific characteristics of different expression classes in the FES datasets: MaEs are more frequent and generally exhibit higher intensity than MEs. This leads the model to focus more on MaEs and generate higher output probabilities for MaE spotting. As a result, more MaEs—including both TPs and FPs—are detected. In contrast, the model generates lower probabilities for ME spotting, resulting in fewer expression proposals but also fewer FPs.

It can also be observed that the overall performance on the SAMM dataset is lower than that on the CAS(ME)$^2$ dataset. The reasons may lie in two main aspects. First, CAS(ME)$^2$ contains very few ME samples, so the ME spotting performance contributes little to the overall performance. This explains why, even though the ME spotting performance of CAS(ME)$^2$ is lower, the overall performance is still maintained at a high level. During both training and inference, the model tends to focus more on MaE spotting, which is relatively easier than ME spotting and has a greater impact on the overall performance. In contrast, for SAMM, since the ground-truth ME samples account for almost one-third of the dataset, ME spotting performance becomes important for the overall performance. Therefore, compared to CAS(ME)$^2$, the model needs to pay more attention to ME spotting on SAMM, which is more challenging in the FES task. Second, the durations of ground-truth MaE samples in CAS(ME)$^2$ are all within a normal range. In contrast, several MaE samples in SAMM abnormally exceed the general duration, making them practically difficult to spot during inference.	

In summary, our proposed model can achieve a more advanced and balanced performance on both datasets, demonstrating the great generalization ability of our model. In addition, our method requires the detection of facial landmarks and the calculation of optical flows to form a head pose-irrelevant graph, making it more robust to variations such as random head angles. 
These merits demonstrate the potential of our method for practical applications (a cross-dataset evaluation on unseen data is provided in Section~\ref{sec:cross_dataset}).

Fig. \ref{videoresult} shows qualitative results on an example video, showcasing both ground-truth expression intervals and detected expression proposals to demonstrate the effectiveness of our model.

\begin{table}[tbp]
\setlength\tabcolsep{2pt}
  \centering
  \caption{Results of cross-dataset validation.}
    \begin{tabular}{llcccccc} 
    \hline
    &&\multicolumn{3}{c}{SAMM-LV}&\multicolumn{3}{c}{CAS(ME)$^2$}\\
    Train&Test&MaE&ME&Overall&MaE&ME&Overall\\
    \hline
    SAMM&CAS(ME)$^2$&-&-&-&0.3497&0.2000&0.3243\\
    CAS(ME)$^2$&SAMM&0.3390&0.1704&0.3008&-&-&-\\
    \hline
    \end{tabular}%
  \label{crossvalidation}%
\end{table}%

\subsection{Cross-dataset validation}
\label{sec:cross_dataset}
To evaluate the robustness and generalization ability of our framework, we conduct cross-dataset validation, where the model is trained on one dataset and directly evaluated on the other without any fine-tuning. Specifically, we train the model on SAMM and evaluate it on CAS(ME)$^2$, and vice versa. This setting simulates a real-world application scenario, as the target-domain distribution differs substantially from the source domain.

As shown in Table~\ref{crossvalidation}, our method achieves strong performance in both cross-dataset directions and consistently outperforms several existing methods, even though it is trained on a completely different dataset. This demonstrates the effectiveness of the proposed CSW-MRO feature and the ability of SpotFormer to capture transferable spatio-temporal cues rather than overfitting to dataset-specific patterns. Notably, despite the limited number of ground-truth ME samples in CAS(ME)$^2$, the model still achieves an ME spotting F1-score of 0.1704 when trained on CAS(ME)$^2$ and evaluated on SAMM, outperforming multiple prior methods. These results confirm the robustness of our framework under significant domain shift and highlight its potential applicability in real-world scenarios where the model must operate on unseen data.

\section{Conclusion}
In this paper, we proposed an efficient framework for FES. First, we proposed an CSW-MRO feature to magnify motion information while alleviating head movement issues. Such an optical flow feature is designed to perceive complete MEs and discern the differences between MaEs and MEs.
Then, we presented SpotFormer, a powerful graph-based Transformer designed to learn multi-scale spatio-temporal relationships from CSW-MRO features for frame-level probability estimation.
In SpotFormer, we proposed an FLGP operation for downsampling facial graph-structured data to achieve multi-scale spatial feature fusion. Moreover, learnable convolution layers are employed for multi-scale temporal feature learning. In addition, we introduced supervised contrastive learning into SpotFormer to enhance the discriminability between different types of expressions at the frame level.
Extensive comparative experiments and ablation studies on the SAMM-LV, CAS(ME)$^2$, and CAS(ME)$^3$ datasets demonstrated the effectiveness of our proposed framework, particularly in ME spotting.

Our model also has some limitations. Performance under conditions with fewer available MEs requires improvement, as evidenced by the experimental results on the CAS(ME)$^2$ dataset. Although our proposed CSW-MRO feature is effective in extracting robust motion features, it is not computationally friendly and highly relies on the accuracy of current optical flow extraction method. Therefore, our method may become less effective under extreme lighting conditions. In future work, we aim to explore an efficient method for solving the data-imbalance problem existing in current datasets and develop an end-to-end spot-then-recognize framework for facial expression analysis.

\section*{Acknowledgments}
This work was partially supported by Innovation Platform for Society 5.0 from Japan Ministry of Education, Culture, Sports, Science and Technology, and JSPS KAKENHI Grant Number JP24K03010.


%



\ifCLASSOPTIONcaptionsoff
  \newpage
\fi

\bibliography{main}

@inproceedings{he2020spotting,
  title={Spotting macro-and micro-expression intervals in long video sequences},
  author={He, Ying and Wang, Su-Jing and Li, Jingting and Yap, Moi Hoon},
  booktitle={Proceedings of the International Conference on Automatic Face and Gesture Recognition},
  pages={742--748},
  year={2020},
  organization={IEEE}
}

@inproceedings{li2021fme,
  title={FME'21: 1st workshop on facial micro-expression: advanced techniques for facial expressions generation and spotting},
  author={Li, Jingting and Yap, Moi Hoon and Cheng, Wen-Huang and See, John and Hong, Xiaopeng and Li, Xiaobai and Wang, Su-Jing},
  booktitle={Proceedings of the International Conference on Multimedia},
  pages={5700--5701},
  year={2021},
    organization={ACM}
}

@article{wang2025micro,
  title={Micro-Expression Key Frame Inference},
  author={Wang, Su-Jing and Miao, Yu-Han and Li, Jingting and Zhou, Ling and Dong, Zizhao and Sun, Mengyi and Fu, Xiaolan},
  journal={IEEE Transactions on Affective Computing},
  year={2025},
  publisher={IEEE}
}

@article{zhang2017facial,
  title={Facial expression recognition based on deep evolutional spatial-temporal networks},
  author={Zhang, Kaihao and Huang, Yongzhen and Du, Yong and Wang, Liang},
  journal={IEEE Transactions on Image Processing},
  volume={26},
  number={9},
  pages={4193--4203},
  year={2017},
  publisher={IEEE}
}

@article{li2024fer,
  title={Fer-former: Multimodal transformer for facial expression recognition},
  author={Li, Yande and Wang, Mingjie and Gong, Minglun and Lu, Yonggang and Liu, Li},
  journal={IEEE Transactions on Multimedia},
  year={2024},
  publisher={IEEE}
}

@article{wang2025visual,
  title={Visual and textual prompts in vllms for enhancing emotion recognition},
  author={Wang, Zhifeng and Zhang, Qixuan and Zhang, Peter and Niu, Wenjia and Zhang, Kaihao and Sankaranarayana, Ramesh and Caldwell, Sabrina and Gedeon, Tom},
  journal={IEEE Transactions on Circuits and Systems for Video Technology},
  year={2025},
  publisher={IEEE}
}

@article{wang2024htnet,
  title={Htnet for micro-expression recognition},
  author={Wang, Zhifeng and Zhang, Kaihao and Luo, Wenhan and Sankaranarayana, Ramesh},
  journal={Neurocomputing},
  volume={602},
  pages={128196},
  year={2024},
  publisher={Elsevier}
}

@inproceedings{see2019megc,
  title={Megc 2019--the second facial micro-expressions grand challenge},
  author={See, John and Yap, Moi Hoon and Li, Jingting and Hong, Xiaopeng and Wang, Su-Jing},
  booktitle={Proceedings of the International Conference on Automatic Face \& Gesture Recognition},
  pages={1--5},
  year={2019},
  organization={IEEE}
}

@inproceedings{yu2024micro,
  title={Micro-Expression Spotting Based on Optical Flow Feature with Boundary Calibration},
  author={Yu, Jun and Zhang, Yaohui and Zhao, Gongpeng and He, Peng and Zhang, Zerui and Cai, Zhongpeng and Liu, Qingsong and Sun, Jianqing and Liang, Jiaen},
  booktitle={Proceedings of International Conference on Multimedia},
  pages={11490--11496},
  year={2024},
organization={ACM}
}

@inproceedings{zhang2024multi,
  title={A Multi-scale Feature Learning Network with Optical Flow Correction for Micro-and Macro-expression Spotting},
  author={Zhang, Zhengye and Zhao, Sirui and Mao, Xinglong and Liu, Shifeng and Wang, Hao and Xu, Tong and Chen, Enhong},
  booktitle={Proceedings of the International Conference on Multimedia},
  pages={11497--11502},
  year={2024},
organization={ACM}
}

@inproceedings{jingting2020megc2020,
  title={Megc2020-the third facial micro-expression grand challenge},
  author={Jingting, LI and Wang, Su-Jing and Yap, Moi Hoon and See, John and Hong, Xiaopeng and Li, Xiaobai},
  booktitle={Proceedings of the International Conference on Automatic Face \& Gesture Recognition},
  pages={777--780},
  year={2020},
  organization={IEEE}
}

@inproceedings{li2022megc2022,
  title={MEGC2022: ACM multimedia 2022 micro-expression grand challenge},
  author={Li, Jingting and Yap, Moi Hoon and Cheng, Wen-Huang and See, John and Hong, Xiaopeng and Li, Xiaobai and Wang, Su-Jing and Davison, Adrian K and Li, Yante and Dong, Zizhao},
  booktitle={Proceedings of the International Conference on Multimedia},
  pages={7170--7174},
  year={2022},
organization={ACM}
}

@inproceedings{davison2023megc2023,
  title={MEGC2023: ACM Multimedia 2023 ME Grand Challenge},
  author={Davison, Adrian K and Li, Jingting and Yap, Moi Hoon and See, John and Cheng, Wen-Huang and Li, Xiaobai and Hong, Xiaopeng and Wang, Su-Jing},
  booktitle={Proceedings of the International Conference on Multimedia},
  pages={9625--9629},
  year={2023},
organization={ACM}
}

@inproceedings{see2024megc2024,
  title={MEGC2024: ACM Multimedia 2024 Facial Micro-Expression Grand Challenge},
  author={See, John and Li, Jingting and Davison, Adrian K and Liong, Gen Bing and Yap, Moi Hoon and Cheng, Wen-Huang and Li, Xiaobai and Hong, Xiaopeng and Wang, Su-Jing},
  booktitle={Proceedings of the International Conference on Multimedia},
  pages={11482--11483},
  year={2024},
organization={ACM}
}

@inproceedings{transformer,
author = {Vaswani, Ashish and Shazeer, Noam and Parmar, Niki and Uszkoreit, Jakob and Jones, Llion and Gomez, Aidan N. and Kaiser, \L{}ukasz and Polosukhin, Illia},
title = {Attention is all you need},
year = {2017},
publisher = {Curran Associates Inc.},
booktitle = {Advances in Neural Information Processing Systems},
pages = {6000–6010},
numpages = {11},
}

@inproceedings{khosla2020supervised,
  title={Supervised contrastive learning},
  author={Khosla, Prannay and Teterwak, Piotr and Wang, Chen and Sarna, Aaron and Tian, Yonglong and Isola, Phillip and Maschinot, Aaron and Liu, Ce and Krishnan, Dilip},
  booktitle={Advances in Neural Information Processing Systems},
  volume={33},
  pages={18661--18673},
  year={2020}
}

@inproceedings{gan2020optical,
  title={Optical strain based macro-and micro-expression sequence spotting in long video},
  author={Gan, YS and Liong, ST and Zheng, D and Li, S and Bin, C},
  booktitle={Proceedings of the International Conference on Automatic Face and Gesture Recognition},
  year={2020},
  organization={IEEE}
}

@inproceedings{zhang2020spatio,
  title={Spatio-temporal fusion for macro-and micro-expression spotting in long video sequences},
  author={Zhang, Li-Wei and Li, Jingting and Wang, Su-Jing and Duan, Xian-Hua and Yan, Wen-Jing and Xie, Hai-Yong and Huang, Shu-Cheng},
  booktitle={Proceedings of the International Conference on Automatic Face and Gesture Recognition},
  pages={734--741},
  year={2020},
  organization={IEEE}
}

@inproceedings{yuhong2021research,
  title={Research on micro-expression spotting method based on optical flow features},
  author={Yuhong, He},
  booktitle={Proceedings of the International Conference on Multimedia},
  pages={4803--4807},
  year={2021},
  organization={ACM}
}

@inproceedings{zhao2022rethinking,
  title={Rethinking Optical Flow Methods for Micro-Expression Spotting},
  author={Zhao, Yuan and Tong, Xin and Zhu, Zichong and Sheng, Jianda and Dai, Lei and Xu, Lingling and Xia, Xuehai and Jiang, Yu and Li, Jiao},
  booktitle={Proceedings of the International Conference on Multimedia},
  pages={7175--7179},
  year={2022},
  organization={ACM}
}

@inproceedings{verburg2019micro,
  title={Micro-expression detection in long videos using optical flow and recurrent neural networks},
  author={Verburg, Michiel and Menkovski, Vlado},
  booktitle={Proceedings of the International Conference on Automatic Face and Gesture Recognition},
  pages={1--6},
  year={2019},
  organization={IEEE}
}

@inproceedings{pan2020local,
  title={Local bilinear convolutional neural network for spotting macro-and micro-expression intervals in long video sequences},
  author={Pan, Hang and Xie, Lun and Wang, Zhiliang},
  booktitle={Proceedings of the International Conference on Automatic Face and Gesture Recognition},
  pages={749--753},
  year={2020},
  organization={IEEE}
}

@article{wang2021mesnet,
  title={{MESNet}: A convolutional neural network for spotting multi-scale micro-expression intervals in long videos},
  author={Wang, Su-Jing and He, Ying and Li, Jingting and Fu, Xiaolan},
  journal={IEEE Transactions on Image Processing},
  volume={30},
  pages={3956--3969},
  year={2021},
  publisher={IEEE}
}

@inproceedings{liong2021shallow,
  title={Shallow optical flow three-stream CNN for macro-and micro-expression spotting from long videos},
  author={Liong, Gen-Bing and See, John and Wong, Lai-Kuan},
  booktitle={Proceedings of the International Conference on Image Processing},
  pages={2643--2647},
  year={2021},
  organization={IEEE}
}

@inproceedings{yap20223d,
  title={{3D-CNN} for facial micro-and macro-expression spotting on long video sequences using temporal oriented reference frame},
  author={Yap, Chuin Hong and Yap, Moi Hoon and Davison, Adrian and Kendrick, Connah and Li, Jingting and Wang, Su-Jing and Cunningham, Ryan},
  booktitle={Proceedings of the International Conference on Multimedia},
  pages={7016--7020},
  year={2022},
  organization={ACM}
}

@inproceedings{yang2021facial,
  title={Facial action unit-based deep learning framework for spotting macro-and micro-expressions in long video sequences},
  author={Yang, Bo and Wu, Jianming and Zhou, Zhiguang and Komiya, Megumi and Kishimoto, Koki and Xu, Jianfeng and Nonaka, Keisuke and Horiuchi, Toshiharu and Komorita, Satoshi and Hattori, Gen and others},
  booktitle={Proceedings of the International Conference on Multimedia},
  pages={4794--4798},
  year={2021},
  organization={ACM}
}

@inproceedings{yu2021lssnet,
  title={{LSSNet}: A two-stream convolutional neural network for spotting macro-and micro-expression in long videos},
  author={Yu, Wang-Wang and Jiang, Jingwen and Li, Yong-Jie},
  booktitle={Proceedings of the International Conference on Multimedia},
  pages={4745--4749},
  year={2021},
  organization={ACM}
}

@inproceedings{liong2022mtsn,
  title={{MTSN}: A Multi-Temporal Stream Network for Spotting Facial Macro-and Micro-Expression with Hard and Soft Pseudo-labels},
  author={Liong, Gen Bing and Liong, Sze-Teng and See, John and Chan, Chee-Seng},
  booktitle={Proceedings of the Workshop on Facial Micro-Expression: Advanced Techniques for Multi-Modal Facial Expression Analysis},
  pages={3--10},
  year={2022},
  organization={ACM}
}

@inproceedings{leng2022abpn,
  title={Abpn: Apex and boundary perception network for micro-and macro-expression spotting},
  author={Leng, Wenhao and Zhao, Sirui and Zhang, Yiming and Liu, Shiifeng and Mao, Xinglong and Wang, Hao and Xu, Tong and Chen, Enhong},
  booktitle={Proceedings of the International Conference on Multimedia},
  pages={7160--7164},
  year={2022},
  organization={ACM}
}

@inproceedings{yin2023aware,
  title={{AU}-aware graph convolutional network for Macroand Micro-expression spotting},
  author={Yin, Shukang and Wu, Shiwei and Xu, Tong and Liu, Shifeng and Zhao, Sirui and Chen, Enhong},
  booktitle={Proceedings of the International Conference on Multimedia and Expo},
  pages={228--233},
  year={2023},
  organization={IEEE}
}

@article{guo2021magnitude,
  title={A magnitude and angle combined optical flow feature for microexpression spotting},
  author={Guo, Yifei and Li, Bing and Ben, Xianye and Ren, Yi and Zhang, Junping and Yan, Rui and Li, Yujun},
  journal={IEEE MultiMedia},
  volume={28},
  number={2},
  pages={29--39},
  year={2021},
  publisher={IEEE}
}

@article{zhang2018smeconvnet,
  title={{SMEConvNet}: A convolutional neural network for spotting spontaneous facial micro-expression from long videos},
  author={Zhang, Zhihao and Chen, Tong and Meng, Hongying and Liu, Guangyuan and Fu, Xiaolan},
  journal={IEEE Access},
  volume={6},
  pages={71143--71151},
  year={2018},
  publisher={IEEE}
}

@inproceedings{
    loshchilov2017decoupled,
    title={Decoupled Weight Decay Regularization},
    author={Ilya Loshchilov and Frank Hutter},
    booktitle={Proceedings of the International Conference on Learning Representations},
    year={2019}
}

@book{ekman2009telling,
  title={Telling lies: Clues to deceit in the marketplace, politics, and marriage (revised edition)},
  author={Ekman, Paul},
  year={2009},
  publisher={WW Norton \& Company}
}

@article{endres2009micro,
  title={Micro-expression recognition training in medical students: a pilot study},
  author={Endres, Jennifer and Laidlaw, Anita},
  journal={BMC Medical Education},
  volume={9},
  number={1},
  pages={1--6},
  year={2009},
  publisher={BioMed Central}
}

@article{o2009police,
  title={Police lie detection accuracy: The effect of lie scenario.},
  author={O'sullivan, Maureen and Frank, Mark G and Hurley, Carolyn M and Tiwana, Jaspreet},
  journal={Law and Human Behavior},
  volume={33},
  number={6},
  pages={530},
  year={2009},
  publisher={Springer}
}

@article{nummenmaaekman,
  title={Emotions Revealed: Recognizing faces and feelings to improve communication and emotional life},
  author={P. Ekman},
  journal={Holt Paperback},
  volume={128},
  number={8},
  pages={140-140},
  year={2003},
}

@article{corneanu2016survey,
  title={Survey on {RGB}, {3D}, thermal, and multimodal approaches for facial expression recognition: History, trends, and affect-related applications},
  author={Corneanu, Ciprian Adrian and Sim{\'o}n, Marc Oliu and Cohn, Jeffrey F and Guerrero, Sergio Escalera},
  journal={IEEE Transactions on Pattern Analysis and Machine Intelligence},
  volume={38},
  number={8},
  pages={1548--1568},
  year={2016},
  publisher={IEEE}
}

@article{yan2013fast,
  title={How fast are the leaked facial expressions: The duration of micro-expressions},
  author={Yan, Wen-Jing and Wu, Qi and Liang, Jing and Chen, Yu-Hsin and Fu, Xiaolan},
  journal={Journal of Nonverbal Behavior},
  volume={37},
  pages={217--230},
  year={2013},
  publisher={Springer}
}

@article{bhushan2015study,
  title={Study of facial micro-expressions in psychology},
  author={Bhushan, Braj},
  journal={Understanding Facial Expressions in Communication: Cross-cultural and Multidisciplinary Perspectives},
  pages={265--286},
  year={2015},
  publisher={Springer}
}

@inproceedings{moilanen2014spotting,
  title={Spotting rapid facial movements from videos using appearance-based feature difference analysis},
  author={Moilanen, Antti and Zhao, Guoying and Pietik{\"a}inen, Matti},
  booktitle={Proceedings of the International Conference on Pattern Recognition},
  pages={1722--1727},
  year={2014},
  organization={IEEE}
}

@inproceedings{davison2015micro,
  title={Micro-facial movement detection using individualised baselines and histogram-based descriptors},
  author={Davison, Adrian K and Yap, Moi Hoon and Lansley, Cliff},
  booktitle={Proceedings of the International Conference on Systems, Man, and Cybernetics},
  pages={1864--1869},
  year={2015},
  organization={IEEE}
}

@article{wang2017main,
  title={A main directional maximal difference analysis for spotting facial movements from long-term videos},
  author={Wang, Su-Jing and Wu, Shuhang and Qian, Xingsheng and Li, Jingxiu and Fu, Xiaolan},
  journal={Neurocomputing},
  volume={230},
  pages={382--389},
  year={2017},
  publisher={Elsevier}
}

@article{liu2015main,
  title={A main directional mean optical flow feature for spontaneous micro-expression recognition},
  author={Liu, Yong-Jin and Zhang, Jin-Kai and Yan, Wen-Jing and Wang, Su-Jing and Zhao, Guoying and Fu, Xiaolan},
  journal={IEEE Transactions on Affective Computing},
  volume={7},
  number={4},
  pages={299--310},
  year={2015},
  publisher={IEEE}
}

@inproceedings{chen2020simple,
  title={A simple framework for contrastive learning of visual representations},
  author={Chen, Ting and Kornblith, Simon and Norouzi, Mohammad and Hinton, Geoffrey},
  booktitle={Proceedings of the International Conference on Machine Learning},
  pages={1597--1607},
  year={2020},
  organization={PMLR}
}

@inproceedings{wangunique,
  title={A unique {M}-pattern for micro-expression spotting in long videos},
  author={Wang, Jinxuan and Xu, Shiting and Zhang, Tong},
  booktitle={Proceedings of the International Conference on Learning Representations},
    year={2024}
}

@inproceedings{PFL,
  title={{PyTorch Face Landmark}: A Fast and Accurate Facial Landmark Detector},
  url={https://github.com/cunjian/pytorch_face_landmark},
  author={Cunjian Chen},
  year={2021},
}

@inproceedings{farneback2003two,
  title={Two-frame motion estimation based on polynomial expansion},
  author={Farneb{\"a}ck, Gunnar},
  booktitle={Proceedings of the Scandinavian Conference on Image Analysis},
  pages={363--370},
  year={2003},
  organization={Springer}
}

@inproceedings{lin2017feature,
  title={Feature pyramid networks for object detection},
  author={Lin, Tsung-Yi and Doll{\'a}r, Piotr and Girshick, Ross and He, Kaiming and Hariharan, Bharath and Belongie, Serge},
  booktitle={Proceedings of the Conference on Computer Vision and Pattern Recognition},
  pages={2117--2125},
  year={2017},
  organization={IEEE/CVF}

}

@inproceedings{liu2021swin,
  title={Swin transformer: Hierarchical vision transformer using shifted windows},
  author={Liu, Ze and Lin, Yutong and Cao, Yue and Hu, Han and Wei, Yixuan and Zhang, Zheng and Lin, Stephen and Guo, Baining},
  booktitle={Proceedings of the International Conference on Computer Vision},
  pages={10012--10022},
  year={2021},
  organization={IEEE/CVF}
}

@inproceedings{lin2017focal,
  title={Focal loss for dense object detection},
  author={Lin, Tsung-Yi and Goyal, Priya and Girshick, Ross and He, Kaiming and Doll{\'a}r, Piotr},
  booktitle={Proceedings of the International Conference on Computer Vision},
  pages={2980--2988},
  year={2017},
  organization={IEEE/CVF}
}

@article{qu2017cas,
  title={{CAS(ME)$^2$}: A Database for Spontaneous Macro-Expression and Micro-Expression Spotting and Recognition},
  author={Qu, Fangbing and Wang, Su-Jing and Yan, Wen-Jing and Li, He and Wu, Shuhang and Fu, Xiaolan},
  journal={IEEE Transactions on Affective Computing},
  volume={9},
  number={4},
  pages={424--436},
  year={2017},
  publisher={IEEE}
}

@inproceedings{yap2020samm,
  title={Samm long videos: A spontaneous facial micro-and macro-expressions dataset},
  author={Yap, Chuin Hong and Kendrick, Connah and Yap, Moi Hoon},
  booktitle={Proceedings of the International Conference on Automatic Face and Gesture Recognition},
  pages={771--776},
  year={2020},
  organization={IEEE}
}

@article{bro2014principal,
  title={Principal component analysis},
  author={Bro, Rasmus and Smilde, Age K},
  journal={Analytical Methods},
  volume={6},
  number={9},
  pages={2812--2831},
  year={2014},
  publisher={Royal Society of Chemistry}
}

@inproceedings{xu2021graph,
  title={Graph stacked hourglass networks for {3D} human pose estimation},
  author={Xu, Tianhan and Takano, Wataru},
  booktitle={Proceedings of the Conference on Computer Vision and Pattern Recognition},
  pages={16105--16114},
  year={2021},
  organization={IEEE/CVF}
}

@article{kopper1996experience,
  title={The experience and expression of anger: Relationships with gender, gender role socialization, depression, and mental health functioning.},
  author={Kopper, Beverly A and Epperson, Douglas L},
  journal={Journal of Counseling Psychology},
  volume={43},
  number={2},
  pages={158},
  year={1996},
  publisher={American Psychological Association}
}

@article{facialvr,
    author = {Thies, Justus and Zollh\"{o}fer, Michael and Stamminger, Marc and Theobalt, Christian and Nie\ss{}ner, Matthias},
    title = {{FaceVR}: Real-Time Gaze-Aware Facial Reenactment in Virtual Reality},
    year = {2018},
    issue_date = {April 2018},
    publisher = {Association for Computing Machinery},
    address = {New York, NY, USA},
    volume = {37},
    number = {2},
    pages={1--15},
    issn = {0730-0301},
    journal = {ACM Transactions on Graphics}
}

@inproceedings{fukuda2002facial,
  title={Facial expression of robot face for human-robot mutual communication},
  author={Fukuda, Toshio and Taguri, Jun and Arai, Fumihito and Nakashima, Masakazu and Tachibana, Daisuke and Hasegawa, Yasuhisa},
  booktitle={Proceedings of the International Conference on Robotics and Automation},
  volume={1},
  pages={46--51},
  year={2002},
  organization={IEEE}
}

@article{yu2023lgsnet,
  author={Yu, Wang-Wang and Jiang, Jingwen and Yang, Kai-Fu and Yan, Hong-Mei and Li, Yong-Jie},
  journal={IEEE Transactions on Affective Computing}, 
  title={{LGSN}et: A Two-Stream Network for Micro- and Macro-Expression Spotting With Background Modeling}, 
  year={2023},
  volume={},
  number={},
  pages={1-18},
  doi={10.1109/TAFFC.2023.3266808}}

@article{GUO2023146,
title = {Micro-expression spotting with multi-scale local transformer in long videos},
journal = {Pattern Recognition Letters},
volume = {168},
pages = {146-152},
year = {2023},
issn = {0167-8655},
doi = {https://doi.org/10.1016/j.patrec.2023.03.012},
author = {Xupeng Guo and Xiaobiao Zhang and Lei Li and Zhaoqiang Xia}
}

@inproceedings{zhang2022actionformer,
  title={Actionformer: Localizing moments of actions with transformers},
  author={Zhang, Chen-Lin and Wu, Jianxin and Li, Yin},
  booktitle={Proceedings of the European Conference on Computer Vision},
  pages={492--510},
  year={2022},
  organization={Springer}
}

@InProceedings{Fan_2021_ICCV,
    author    = {Fan, Haoqi and Xiong, Bo and Mangalam, Karttikeya and Li, Yanghao and Yan, Zhicheng and Malik, Jitendra and Feichtenhofer, Christoph},
    title     = {Multiscale Vision Transformers},
    booktitle = {Proceedings of the International Conference on Computer Vision},
    month     = {October},
    year      = {2021},
    pages     = {6824-6835},
    organization={IEEE/CVF}
}

@InProceedings{Arnab_2021_ICCV,
    author    = {Arnab, Anurag and Dehghani, Mostafa and Heigold, Georg and Sun, Chen and Lu\v{c}i\'c, Mario and Schmid, Cordelia},
    title     = {Vi{V}iT: A Video Vision Transformer},
    booktitle = {Proceedings of the International Conference on Computer Vision},
    month     = {October},
    year      = {2021},
    pages     = {6836-6846},
    organization = {IEEE/CVF}
}

@InProceedings{Shao_2023_ICCV,
    author    = {Shao, Jiayi and Wang, Xiaohan and Quan, Ruijie and Zheng, Junjun and Yang, Jiang and Yang, Yi},
    title     = {Action Sensitivity Learning for Temporal Action Localization},
    booktitle = {Proceedings of the International Conference on Computer Vision},
    month     = {October},
    year      = {2023},
    pages     = {13457-13469},
    organization = {IEEE/CVF}
}

@inproceedings{
dosovitskiy2021an,
title={An Image is Worth 16x16 Words: Transformers for Image Recognition at Scale},
author={Alexey Dosovitskiy and Lucas Beyer and Alexander Kolesnikov and Dirk Weissenborn and Xiaohua Zhai and Thomas Unterthiner and Mostafa Dehghani and Matthias Minderer and Georg Heigold and Sylvain Gelly and Jakob Uszkoreit and Neil Houlsby},
booktitle={Proceedings of the International Conference on Learning Representations},
year={2021},
}

@InProceedings{Liu_2022_CVPR,
    author    = {Liu, Ze and Ning, Jia and Cao, Yue and Wei, Yixuan and Zhang, Zheng and Lin, Stephen and Hu, Han},
    title     = {Video Swin Transformer},
    booktitle = {Proceedings of the Conference on Computer Vision and Pattern Recognition},
    month     = {June},
    year      = {2022},
    pages     = {3202-3211},
    organization = {IEEE/CVF}
}

@ARTICLE{9774929,
  author={Li, Jingting and Dong, Zizhao and Lu, Shaoyuan and Wang, Su-Jing and Yan, Wen-Jing and Ma, Yinhuan and Liu, Ye and Huang, Changbing and Fu, Xiaolan},
  journal={IEEE Transactions on Pattern Analysis and Machine Intelligence}, 
  title={CAS(ME)3: A Third Generation Facial Spontaneous Micro-Expression Database With Depth Information and High Ecological Validity}, 
  year={2023},
  volume={45},
  number={3},
  pages={2782-2800},
  doi={10.1109/TPAMI.2022.3174895}}

@inproceedings{zou2025synergistic,
  title={Synergistic spotting and recognition of micro-expression via temporal state transition},
  author={Zou, Bochao and Guo, Zizheng and Qin, Wenfeng and Li, Xin and Wang, Kangsheng and Ma, Huimin},
  booktitle={Proceedings of the International Conference on Acoustics, Speech and Signal Processing},
  year={2025},
  organization={IEEE}
}

@article{li2025could,
  title={Could micro-expressions be quantified? Electromyography gives affirmative evidence},
  author={Li, Jingting and Lu, Shaoyuan and Wang, Yan and Dong, Zizhao and Wang, Su-Jing and Fu, Xiaolan},
  journal={IEEE Transactions on Affective Computing},
  year={2025},
  publisher={IEEE}
}

@InProceedings{Xu_2017_ICCV,
author = {Xu, Huijuan and Das, Abir and Saenko, Kate},
title = {R-C3D: Region Convolutional 3D Network for Temporal Activity Detection},
booktitle = {Proceedings of the International Conference on Computer Vision (ICCV)},
month = {Oct},
year = {2017},
organization = {IEEE}
}

@InProceedings{Chao_2018_CVPR,
author = {Chao, Yu-Wei and Vijayanarasimhan, Sudheendra and Seybold, Bryan and Ross, David A. and Deng, Jia and Sukthankar, Rahul},
title = {Rethinking the Faster R-CNN Architecture for Temporal Action Localization},
booktitle = {Proceedings of the Conference on Computer Vision and Pattern Recognition},
month = {June},
year = {2018},
organization = {IEEE}
}

@InProceedings{Lin_2018_ECCV,
author = {Lin, Tianwei and Zhao, Xu and Su, Haisheng and Wang, Chongjing and Yang, Ming},
title = {BSN: Boundary Sensitive Network for Temporal Action Proposal Generation},
booktitle = {Proceedings of the European Conference on Computer Vision},
month = {September},
year = {2018},
organization = {Springer}
}

@InProceedings{Nag_2023_CVPR,
    author    = {Nag, Sauradip and Zhu, Xiatian and Song, Yi-Zhe and Xiang, Tao},
    title     = {Post-Processing Temporal Action Detection},
    booktitle = {Proceedings of the Conference on Computer Vision and Pattern Recognition},
    month     = {June},
    year      = {2023},
    pages     = {18837-18845},
organization = {IEEE/CVF}
}

@article{yang2020revisiting,
  title={Revisiting anchor mechanisms for temporal action localization},
  author={Yang, Le and Peng, Houwen and Zhang, Dingwen and Fu, Jianlong and Han, Junwei},
  journal={IEEE Transactions on Image Processing},
  volume={29},
  pages={8535--8548},
  year={2020},
  publisher={IEEE}
}

@InProceedings{zhang2021STST,
author = {Zhang, Yuhan and Wu, Bo and Li, Wen and Duan, Lixin and Gan, Chuang},
title = {{STST}: Spatial-Temporal Specialized Transformer for Skeleton-based Action Recognition},
year = {2021},
booktitle = {Proceedings of the International Conference on Multimedia},
pages = {3229–3237},
numpages = {9},
organization = {ACM}
}

@InProceedings{deng2024SpoTGCN,
  author    = {Deng, Yicheng and Hayashi, Hideaki and Nagahara, Hajime},
  title     = {Multi-Scale Spatio-Temporal Graph Convolutional Network for Facial Expression Spotting},
  booktitle      = {Proceedings of the International Conference on Automatic Face and Gesture Recognition},
  year      = {2024},
  organization = {IEEE}
}

@inproceedings{carreira2017quo,
  title={Quo vadis, action recognition? a new model and the kinetics dataset},
  author={Carreira, Joao and Zisserman, Andrew},
  booktitle={Proceedings of the Conference on Computer Vision and Pattern Recognition},
  pages={6299--6308},
  year={2017},
    organization = {IEEE}
}

@inproceedings{pan2021spatio,
  title={Spatio-temporal convolutional attention network for spotting macro-and micro-expression intervals},
  author={Pan, Hang and Xie, Lun and Wang, Zhiliang},
  booktitle={Proceedings of the 1st Workshop on Facial Micro-Expression: Advanced Techniques for Facial Expressions Generation and Spotting},
  pages={25--30},
  year={2021}
}

@inproceedings{yu2022facial,
  title={Facial expression spotting based on optical flow features},
  author={Yu, Jun and Cai, Zhongpeng and Liu, Zepeng and Xie, Guochen and He, Peng},
  booktitle={Proceedings of the International Conference on Multimedia},
  pages={7205--7209},
  year={2022},
  organization={ACM}
}

@inproceedings{qin2023micro,
  title={Micro-expression spotting with face alignment and optical flow},
  author={Qin, Wenfeng and Zou, Bochao and Li, Xin and Wang, Weiping and Ma, HUimin},
  booktitle={Proceedings of the International Conference on Multimedia},
  pages={9501--9505},
  year={2023},
    organization={ACM}

}

@inproceedings{yu2023efficient,
  title={Efficient micro-expression spotting based on main directional mean optical flow feature},
  author={Yu, Jun and Cai, Zhongpeng and Du, Shenshen and Shen, Xiaxin and Wang, Lei and Gao, Fang},
  booktitle={Proceedings of the International Conference on Multimedia},
  pages={9541--9545},
  year={2023},
    organization={ACM}

}

@article{guo2023micro,
  title={Micro-expression spotting with multi-scale local transformer in long videos},
  author={Guo, Xupeng and Zhang, Xiaobiao and Li, Lei and Xia, Zhaoqiang},
  journal={Pattern Recognition Letters},
  volume={168},
  pages={146--152},
  year={2023},
  publisher={Elsevier}
}

@article{yang2026mul,
  title={Mul-FES: A multimodal facial expression spotting method that integrates textual information},
  author={Yang, Henian and Huang, Shucheng and Li, Mingxing},
  journal={Expert Systems with Applications},
  volume={304},
  pages={129332},
  year={2026},
  publisher={Elsevier}
}

@article{yang2025ofct,
  title={Ofct: a micro-expression spotting method fusing optical flow features and category text information},
  author={Yang, Henian and Huang, Shucheng and Li, Mingxing},
  journal={Complex \& Intelligent Systems},
  volume={11},
  number={8},
  pages={373},
  year={2025},
  publisher={Springer}
}

@inproceedings{li2024learning,
  title={Learning interval-aware embedding for macro-and micro-expression spotting},
  author={Li, Xiaodong and Li, Jiajun and Du, Wenchao and Chen, Hu and Yang, Hongyu},
  booktitle={Proceedings of the Asian Conference on Computer Vision},
  pages={337--353},
  year={2024}
}
%

\bibliographystyle{IEEEtran}

\begin{IEEEbiography}[{\includegraphics[width=1in,height=1.25in,clip,keepaspectratio]{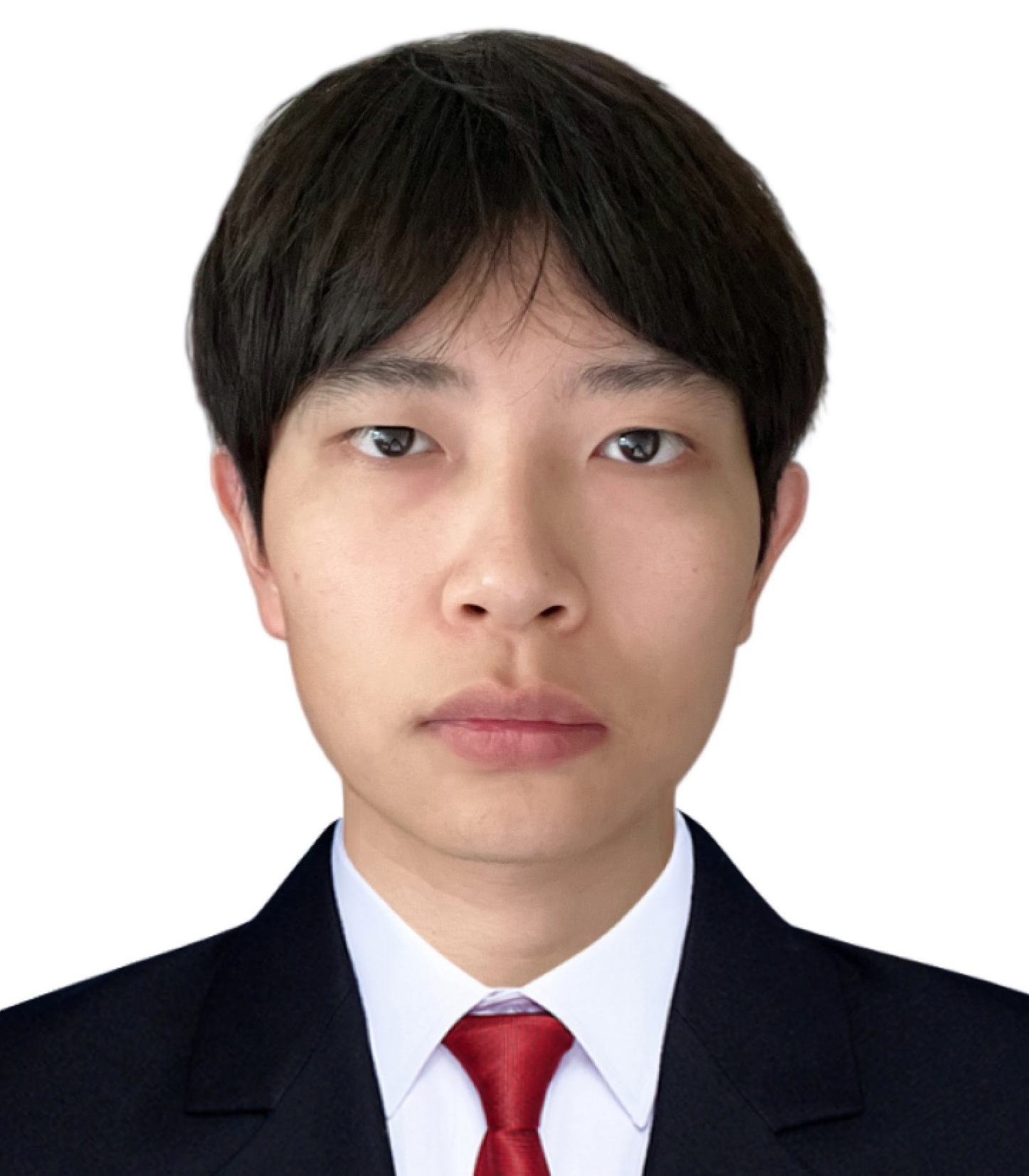}}]{Yicheng Deng}
received the B.Eng. and M.Eng. degrees from Beijing Jiaotong University, Beijing, China, in 2019 and 2022, respectively. He is now pursuing his Ph.D. degree in The University of Osaka. His research interests include facial expression analysis and video understanding.
\end{IEEEbiography}
\begin{IEEEbiography}[{\includegraphics[width=1in,height=1.25in,clip,keepaspectratio]{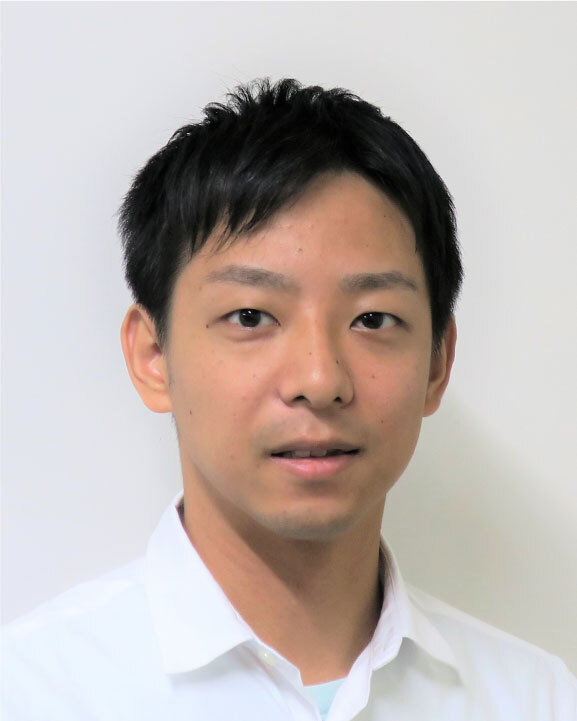}}]{Hideaki Hayashi}
(S' 13--M' 16) received the B.E., M.Eng, and D.Eng. degrees from Hiroshima University, Hiroshima, Japan, in 2012, 2014, and 2016 respectively. He was a Research Fellow of the Japan Society for the Promotion of Science from 2015 to 2017 and an assistant professor in Department of Advanced Information Technology, Kyushu University from 2017 to 2022. He is currently an associate professor with the D3 Center, The University of Osaka. His research interests focus on neural networks, machine learning, and medical data analysis. 
\end{IEEEbiography}
\begin{IEEEbiography}[{\includegraphics[width=1in,height=1.25in,clip,keepaspectratio]{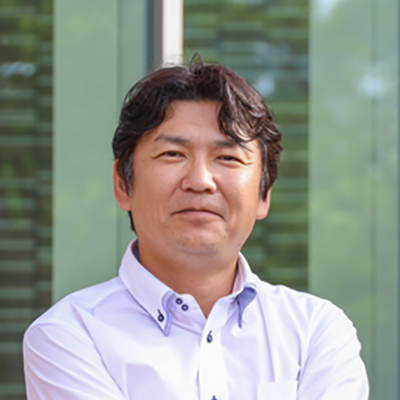}}]{Hajime Nagahara} received Ph.D. degree in system engineering from The University of Osaka in 2001. He was a research associate of the Japan Society for the Promotion of Science from 2001 to 2003. He was an assistant professor at the Graduate School of Engineering Science, The University of Osaka, Japan from 2003 to 2010.  He was an associate professor in Faculty of Information Science and Electrical Engineering at Kyushu University from 2010 to 2017. He was a professor at Institute for Datability Science, The University of Osaka, from 2017 to 2024.  
He was a visiting associate professor at CREA University of Picardie Jules Verns, France, in 2005. He was a visiting researcher at Columbia University in 2007-2008 and 2016-2017. He has been a professor at D3 Center, The University of Osaka, since 2024.
Computational photography and computer vision are his research areas. He received IPSJ Nagao Special Researcher Award in 2012, ICCP2016 Best Paper Runners-up, and SSII Takagi Award in 2016. He is the Program Chair for ICCP2019, the General Chair for ACCV2026, an Associate Editor for IEEE Transactions on Computational Imaging from 2019 to 2022, an Associate Editor for IEEE Transactions on Pattern Analysis and Machine Intelligence from 2023, and the Director of the Information Processing Society of Japan from 2022 to 2024.
\end{IEEEbiography}

\end{document}